\documentclass{article}

\usepackage{mylatexstyle}

\usepackage[utf8]{inputenc}
\usepackage[T1]{fontenc}
\usepackage[left=1in, right=1in, top=1in, bottom=1in]{geometry}

\usepackage{hyperref}
\usepackage{url}
\usepackage{booktabs}
\usepackage{amsfonts,amsmath,amssymb,amsthm}
\usepackage{nicefrac}
\usepackage{microtype}
\usepackage[table, dvipsnames]{xcolor}
\usepackage{tcolorbox}
\usepackage{algorithm}
\usepackage{algpseudocode}
\usepackage{dirtree}
\usepackage{graphicx}
\usepackage{tikz}
\usetikzlibrary{arrows.meta,positioning,shapes.geometric,calc}
\usepackage{wrapfig}

\tikzset{
  flow/.style={draw=Dark,fill=TanVlt,rounded corners=3pt,very thick,align=center,inner xsep=8pt,inner ysep=7pt},
  gate/.style={draw=Dark,fill=Tan,rounded corners=3pt,very thick,align=center,inner xsep=8pt,inner ysep=7pt,font=\bfseries},
  arrow/.style={-Latex,thick,draw=Dark},
  dashedarrow/.style={-Latex,thick,dashed,draw=Dark},
  smallcard/.style={draw=Dark,fill=TanVlt,rounded corners=4pt,thick,align=left,inner xsep=7pt,inner ysep=6pt,text width=0.28\linewidth}
}

\definecolor{BgCream}{HTML}{F2ECD8}
\definecolor{Tan}{HTML}{C8B988}
\definecolor{TanLt}{HTML}{D9CDA8}
\definecolor{TanVlt}{HTML}{E8DEBE}
\definecolor{Dark}{HTML}{2A2418}
\definecolor{Blue}{HTML}{2F6DC2}
\definecolor{Green}{HTML}{62A55A}
\definecolor{Red}{HTML}{D9483D}
\definecolor{Orange}{HTML}{E0822F}
\definecolor{Purple}{HTML}{9852AE}
\definecolor{Brown}{HTML}{8B6A3F}

\definecolor{codetext}{HTML}{BA4F1D}

\newtcbox{\code}{
  on line,
  boxrule=0pt,
  boxsep=0pt,
  top=2pt, bottom=2pt, left=4pt, right=4pt,
  colback=gray!15,
  coltext=codetext,
  arc=3pt,
  fontupper=\ttfamily
}

\usepackage{listings}
\usepackage{listingsutf8}
\usepackage{enumitem}

\lstdefinestyle{lean}{
  basicstyle=\ttfamily\small,
  backgroundcolor=\color{gray!10},
  frame=single,
  framesep=6pt,
  rulecolor=\color{gray!40},
  breaklines=true,
  columns=fullflexible,
  keepspaces=true,
  showstringspaces=false,
  commentstyle=\color{gray!70!black},
  keywordstyle=\color{blue!70!black}\bfseries,
  keywordstyle=[2]\color{codetext},
  stringstyle=\color{purple!70!black},
  morekeywords={lemma,theorem,def,by},
  morekeywords=[2]{statement,proof,title,latexEnv},
  morecomment=[l]{--},
  morestring=[b]",
}

\usepackage[capitalize]{cleveref}
\crefname{equation}{Eq.}{Eqs.}
\Crefname{equation}{Eq.}{Eqs.}

\usepackage{aliascnt}

\theoremstyle{plain}

\newaliascnt{proposition}{theorem}

\aliascntresetthe{proposition}
\crefname{proposition}{Proposition}{Propositions}

\newaliascnt{lemma}{theorem}

\aliascntresetthe{lemma}
\crefname{lemma}{Lemma}{Lemmas}

\newaliascnt{corollary}{theorem}

\aliascntresetthe{corollary}
\crefname{corollary}{Corollary}{Corollaries}

\theoremstyle{definition}
\newaliascnt{definition}{theorem}

\aliascntresetthe{definition}
\crefname{definition}{Definition}{Definitions}

\newaliascnt{assumption}{theorem}

\aliascntresetthe{assumption}
\crefname{assumption}{Assumption}{Assumptions}

\theoremstyle{remark}
\newaliascnt{remark}{theorem}

\aliascntresetthe{remark}
\crefname{remark}{Remark}{Remarks}

\providecommand{\agent}[1]{\textsc{#1}}

\usepackage{xspace}
\newcommand{\agentname}{LeanMarathon\xspace}

\usepackage{eso-pic}
\AddToShipoutPictureFG*{%
  \AtPageUpperLeft{%
    \raisebox{-2cm}{%
      \hspace{2.5cm}
      \includegraphics[width=1.5cm]{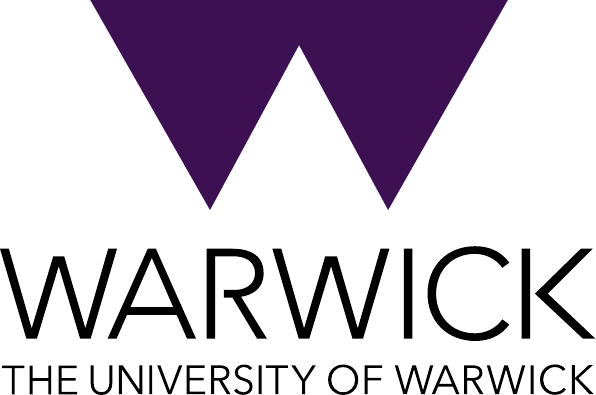}%
      \hspace{0.35cm}%
      \includegraphics[width=2cm]{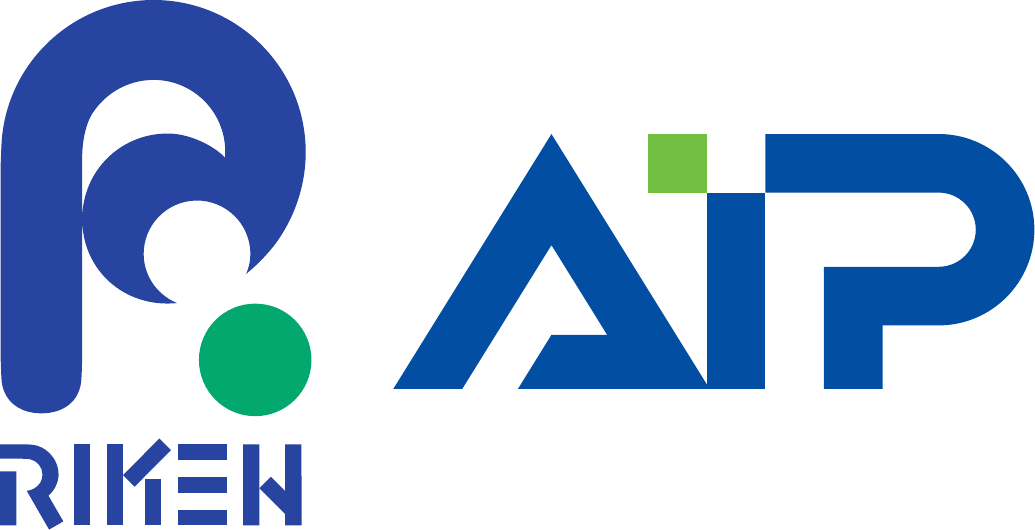}%
      \hspace{0.3cm}%
      \includegraphics[width=1cm]{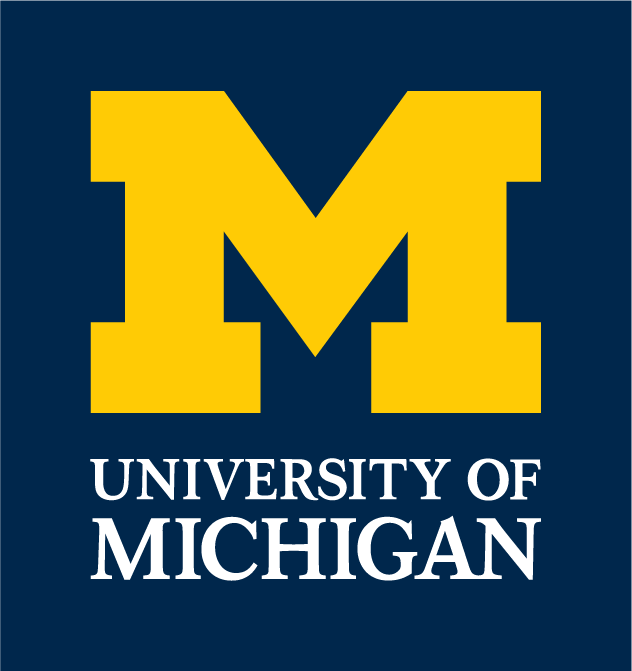}%
      \hspace{0.35cm}%
      \includegraphics[width=1cm]{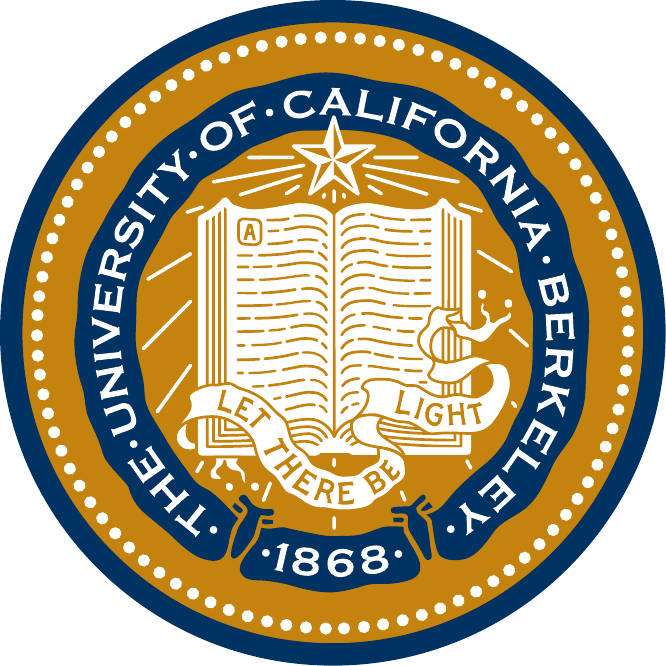}%
      \hspace{0.3cm}
      \includegraphics[width=1cm]{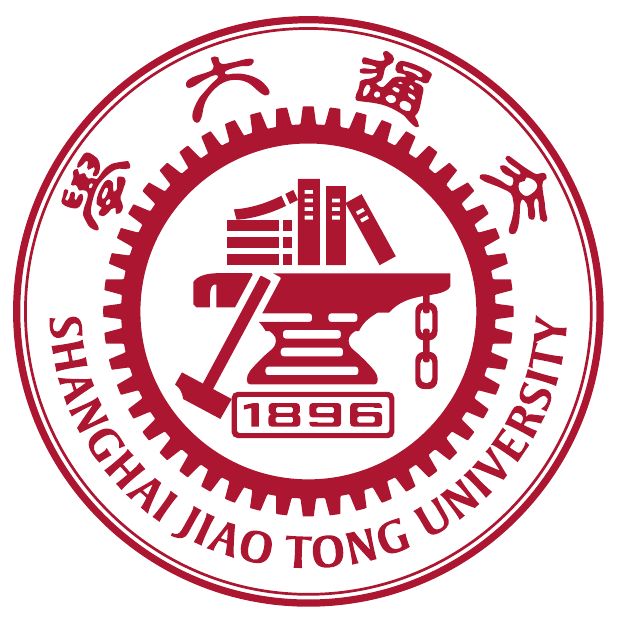}%
    }%
  }%
}

\AddToShipoutPictureFG*{%
  \AtPageUpperLeft{%
    \raisebox{-2.2cm}{%
      \hspace*{\dimexpr 1in+\oddsidemargin\relax}%
      \makebox[\textwidth][l]{\rule{\textwidth}{0.5pt}}
    }%
  }%
}

\title{LeanMarathon: Toward Reliable AI Co-Mathematicians through Long-Horizon Lean Autoformalization}
\author
{
     Yuanhe Zhang\thanks{Department of Statistics, University of Warwick, UK; also Center for Advanced Intelligence Project, RIKEN, Japan. Email: {\tt yuanhe.zhang@warwick.ac.uk}} 
     ~~~
     Yuekai Sun\thanks{Department of Statistics, University of Michigan, USA. Email: {\tt yuekai@umich.edu}}
     ~~~
     Taiji Suzuki\thanks{Department of Mathematical Informatics, The University of Tokyo; also Center for Advanced Intelligence Project, RIKEN, Japan. Email: {\tt taiji@mist.i.u-tokyo.ac.jp}}
     ~~~
     Jason D. Lee\thanks{Department of Electrical Engineering and Computer Sciences, also Department of Statistics, University of California, Berkeley, USA. Email: {\tt jasondlee@berkeley.edu}}
     ~~~
     Fanghui Liu\thanks{School of Mathematical Sciences, Institute of Natural Sciences and MOE-LSC, Shanghai Jiao Tong University, China. Email: {\tt fanghui.liu@sjtu.edu.cn} (Corresponding author)}
}
\date{}

\begin{document}

\maketitle
\begin{figure}[h!]
    \centering
    \vspace{-1cm}
    \includegraphics[width=0.45\linewidth]{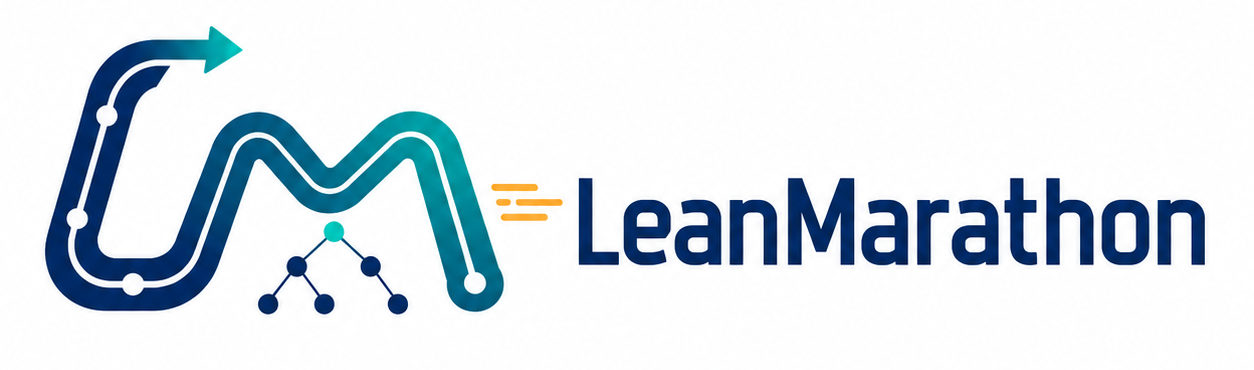}
    \vspace{-0.5cm}
\end{figure}

\begin{abstract}
Long-horizon autoformalization of research mathematics fails not only at hard lemmas, but at scale: statements drift, dependencies tangle, context decays, and local repairs corrupt distant work. We present \agentname, a multi-agent harness for reliable research-level Lean autoformalization. Its core abstraction is an evolving blueprint: a Lean file that serves simultaneously as formal proof skeleton, natural-language proof graph, and shared system of record. Four contract-scoped agents construct, audit, prove, and repair this blueprint. These agents are coordinated by a two-stage orchestrator that first stabilizes target fidelity through adversarial review and then discharges the proof directed acyclic graph (DAG) from its dynamic leaves upward in parallel CI-gated rounds.
\agentname turns one brittle multi-hour run into many local, recoverable, parallel transactions. 
We evaluate \agentname on two recent research papers spanning four Erd\H{o}s problems (\#1051, \#1196, \#164, \#1217). Across three autonomous runs, it formalizes all seven target theorems with no \code{sorry}, proving 258 lemmas and theorems.
These results show that reliable AI co-mathematics requires not only stronger provers, but durable harnesses that preserve target fidelity across long mathematical developments. The code can be found at \url{https://github.com/YuanheZ/LeanMarathon}.
\end{abstract}

\section{Introduction}
\label{sec:intro}

AI-assisted mathematics can be organized into three interlocking stages, as emphasized by~\citet{tao2026threecomponents}: proof \emph{generation} large language models (LLMs), \emph{verification} via Lean 4, and \emph{digestion} by human. Generation by LLMs has advanced fast, such as Aletheia~\citep{aletheia,zheng2026comath} and GPT-5~\citep{bubeck2025gpt5science}. These agents can produce long natural-language proofs and solve open problems, subject to human verification. Verification, the stage that turns such a natural language proof into a machine-checked artifact, has kept pace only on isolated goals: a prover discharges one or multiple Lean~4 lemmas at a time. Verifying an \emph{entire research paper} remains largely open: every definition, lemma, and theorem must be formalized so the whole argument type-checks with no \code{sorry}. 

The difficulty is therefore not merely the formalization of one hard lemma. A research paper may require hundreds of mutually dependent declarations, and these declarations must remain coherent while an autonomous system repeatedly edits, checks, and repairs a growing Lean development. A local change to a definition can invalidate distant proofs; a misformalized intermediate lemma can make downstream work formally correct but mathematically irrelevant; and a seemingly successful repair can silently move the formal proof away from the intended theorem. Long-running autoformalization thus fails in ways that resemble software-engineering failure as much as theorem-proving failure: stale context, dependency tangles, statement drift, and repairs whose effects are hard to localize.

Automating this verification stage is important for two reasons. First, machine-checked formalization, e.g., Lean 4, is the strongest available guarantee that an AI-discovered proof is correct rather than merely plausible. It is therefore a prerequisite for making AI-assisted mathematics reliable enough to {\bf \em accelerate research at scale}. Second, formalization can also support proof digestion. Since a Lean proof records a proof at a very fine level of detail, it can be translated back into natural language at multiple resolutions: a high-level overview for orientation, a structured proof outline for learning, and a fully detailed derivation for verification. This suggests a future form of mathematical textbook in which the same formal artifact supports both machine checking and human-facing explanations at different levels of granularity.

For research-level formalization, the core bottleneck is formalization shifts. The limiting factor is not (just) the model's capability on a single goal but its \emph{agent durability}: whether an autonomous system stays coherent across a multi-hour run, preserves the intended target, calibrates the failure state, and keeps one wrong decision from corrupting the rest of the proof.
This differs significantly from textbook formalization~\citep{wang2026m2f,gloeckle2026automatic} that provides the agent with a fine-grained blueprint of the full reasoning pipeline then the agent’s core task is to translate this existing blueprint into formal Lean code, rather than to discover the logical structure of the reasoning itself. In research-level formalization, no such blueprint is available in advance. The only fixed anchors are the terminal target theorems specified by human researchers. The source proof, whether written by humans or generated with AI assistance, may contain noise, implicit steps, gaps, or even errors.

This absence of a trusted fine-grained proof plan creates two pervasive failure modes. The first is \emph{goal drift}, where the agent’s intermediate reasoning gradually deviates from the logical path required to prove the target theorem, resulting in a formally correct but irrelevant reasoning graph; The second is \emph{lost-in-the-middle}, where the agent becomes trapped in an exponentially growing space of unproductive subproblems, unable to navigate back to the core target or prioritize high-impact intermediate steps.
Both failure modes are relatively muted in textbook settings where the proof graph is supplied, but they become central in research settings where the system must discover, maintain, and repair the graph itself.

A monolithic agent asked to read the paper, design the formal skeleton, prove the lemmas, diagnose failures, and repair its own mistakes is fragile. A single defect such as missed hypothesis, drifted statement, or confident but wrong repair can invalidate hours of downstream work with high risk of context rot. The question addressed in this paper is therefore not simply how to make an LLM prove a Lean goal, but how to design a multi-agent harness that makes long-horizon Lean autoformalization of research mathematics legible, recoverable, and resistant to drift. More generally, our work explores the possibility of autoformalization of research mathematics via harness design if the related Lean library (e.g., Mathlib) is sufficient and complete.

\begin{figure}[t]
    \centering
    \includegraphics[width=\linewidth]{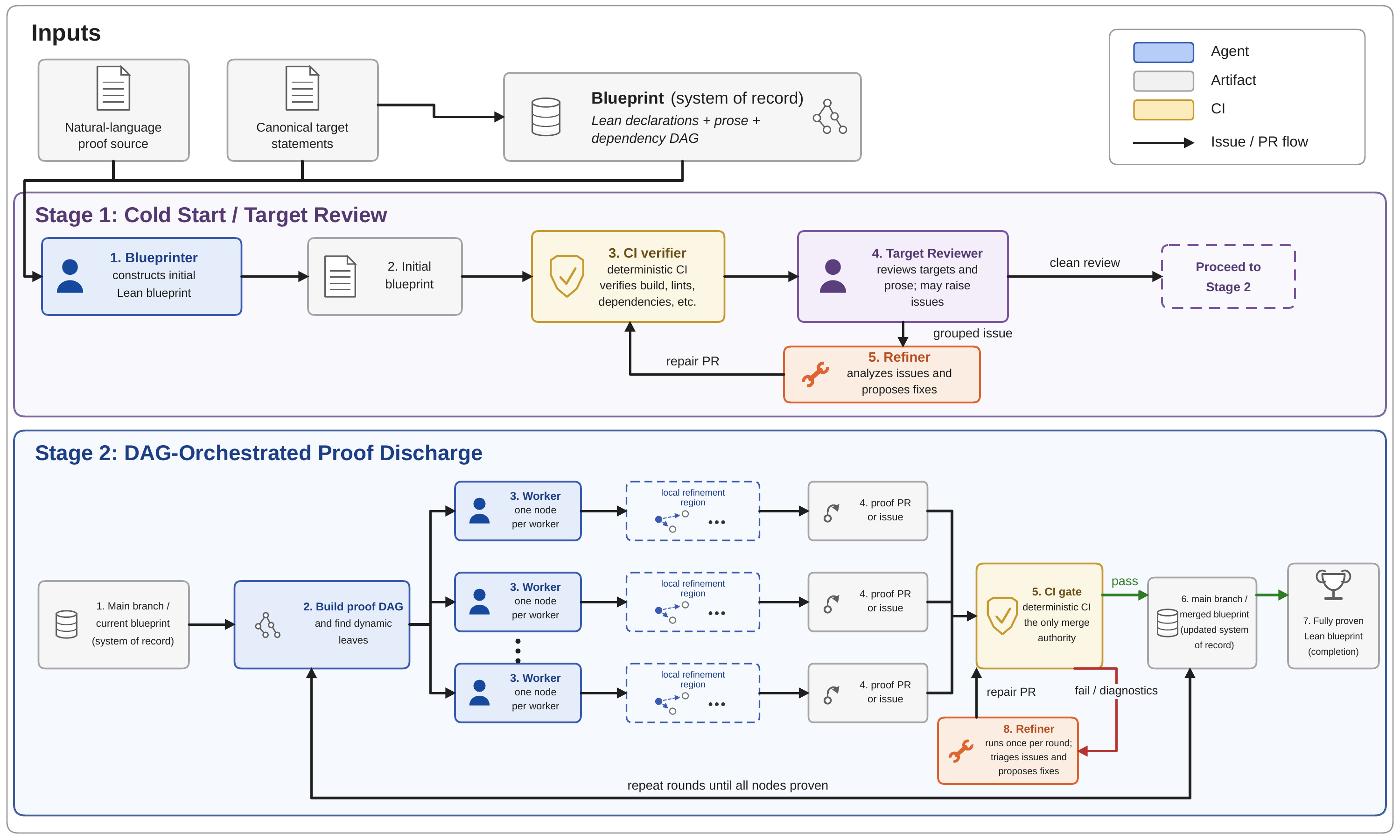}
    \caption{Overview of \agentname.}
    \label{fig-pipeline}
\end{figure}

\subsection{Contributions}

We propose \agentname, a multi-agent harness for long-horizon Lean autoformalization of research mathematics, exemplified by Erd\H{o}s problems (\#1051, \#1196, \#164, \#1217) from two recent papers \citep{erdosGraham2026,primitiveChains2026}. At its center is the \emph{blueprint}: a single Lean file that is at once a formal proof skeleton and a natural-language proof graph. Each \code{lemma} or \code{theorem} is a \emph{node} pairing a Lean type with the LaTeX statement and proof it formalizes; when one node's proof invokes another, it induces a directed dependency edge. These nodes and edges form the \emph{proof directed acyclic graph (DAG)} that every agent reads, extends, and repairs (definitions are shared global context, not nodes). The formalization is complete when every node carries a Lean proof and the file type-checks with no \code{sorry}.
We expect the proof DAG as well as harness design for long-horizon autoformalization follows four principles:
\begin{itemize}
  \item \textbf{Decompose with dynamic proof DAG.} The initial decomposition is uncertain, so the system never freezes it. It generates the DAG and then lets it evolve, splitting an over-large node or repairing a misformalized one, until each piece is small enough for one agent to discharge.
  \item \textbf{External or deterministic verification.} No agent judges its own output.
  The progress is decided by the the deterministic verifier and external agents, not by an agent's self-assessment, which avoids potential formalization drifts.
  \item \textbf{Restrict tools scope.} Each agent is expected to edit only a bounded region. Constraining the action space calibrates the agent's trace, so the worst outcome of a mistake is a rejected patch rather than corruption of another agent's state.
  \item \textbf{Informalize while formalizing.} Each node keeps its LaTeX prose beside its Lean type. This keeps the artifact human-readable and also guards against drift: the verifier enforces parity between the prose dependency graph and the elaborator's, so the natural-language and formal graphs cannot silently diverge.
\end{itemize}

\cref{fig-pipeline} shows how \agentname instantiates these principles. Four contract-scoped agents share the blueprint, each with a narrow input, a single output, and a bounded edit scope. The \agent{Blueprinter} reads the source proof and the canonical target statements and writes the initial skeleton, with every statement type-checking and every proof body a \code{sorry} placeholder. The \agent{Target-Reviewer} audits that skeleton for fidelity, checking that each Lean type states the theorem the paper intends, and files an issue on any mismatch. The \agent{Worker} discharges one node behind a quality gate, proving its body inside a frozen local region. The \agent{Refiner} collects open issues and repairs multi-node defects in one pull request (PR) per round. 

These agents are coordinated by a two-stage orchestrator. Stage 1 runs an adversarial review loop between the \agent{Target-Reviewer} and \agent{Refiner} until the target statements are certified as faithful. Stage 2 repeatedly extracts the current proof DAG, identifies dynamic leaves whose dependencies have already been proved, and assigns them to Workers in parallel. All pull requests pass through a deterministic CI gate before reaching \code{main}; passing non-conflicting PRs are squash-merged independently as their checks complete. The contributions of this paper can be summarized as below.

\begin{itemize}
\item We identify \textbf{agent durability} as the central bottleneck in research-level autoformalization, and characterize three failure modes any long-running agent exhibits. \emph{Coherence loss}: the task is globally coupled, like Sudoku, where every step must respect the whole proof, so a monolithic agent turns myopic and robs one part of the proof to patch another. \emph{Self-evaluation bias}: asked to evaluate its own output, the agent appears to be overly confident. \emph{Irreversibility}: once the work drifts from the target, the agent cannot recover it, and the damage compounds across the rest of the proof. A single agent will always make such errors, so the design goal is not an infallible agent but a system of fallible agents in which no one error spreads: \textbf{fault containment}. We realize it through four contract-scoped agents and a two-stage orchestrator that separates adversarial target review from parallel proof discharge, turning one brittle multi-day run into many short, recoverable, parallel ones.
\item Our core formulation is the \textbf{dynamic proof DAG}: one blueprint that is at once a formal Lean skeleton, a natural-language proof graph, and the shared system of record. The harness never freezes the initial decomposition; instead it grows and repairs the DAG and discharges it from its dynamic leaves upward in parallel. Holding this moving graph coherent is the job of the whole harness, enforced concretely by \textbf{deterministic CI gate} and encoded into the contracts of agents.
\item We evaluate \agentname on \textbf{research-level Erd\H{o}s problems} for autoformalization. Across two 2026 papers spanning four Erd\H{o}s problems~\citep{erdosGraham2026,primitiveChains2026}, \agentname formalizes all seven target theorems with no \code{sorry}, proving 258 lemmas and theorems in total. A commercial single-agent baseline~\citep{aristotle} fails on both papers after tens of hours. \agentname localizes failures to individual nodes and keeps incorrect early formalizations from silently consuming downstream prover compute.
\end{itemize}

\subsection{Related Work}
\label{sec:related}

\paragraph{Autoformalization and neural theorem proving.}
Early systems prompt an LLM to translate statements~\citep{wu2022autoformalization} or to draft informal proofs into formal sketches~\citep{jiang2023dsp}, and later work scales paired data and Lean-specific translators~\citep{ying2024leanworkbook,gao2025herald}. In parallel, neural provers search for proofs: tactic models trained on proof data~\citep{polu2020gptf,han2022pact}, tree search~\citep{lample2022htps}, premise retrieval~\citep{yang2023leandojo}, interleaved informal reasoning~\citep{lin2024leanstar}, reinforcement learning from proof-assistant feedback and subgoal decomposition~\citep{xin2024deepseekproverv15,ren2025deepseekproverv2}, self-correction from compiler errors~\citep{lin2025goedelproverv2}, and large reasoning provers~\citep{wang2025kiminaprover,chen2025seedprover}. The targets which can be formalized by these systems are at most IMO-level.

\paragraph{AI for research-level mathematics.}
Recent systems push AI from competition problems toward open research questions. AlphaProof reached olympiad-medal level by proving in Lean with reinforcement learning~\citep{hubert2026alphaproof}, while AlphaGeometry~\citep{trinh2024alphageometry} and AlphaEvolve~\citep{novikov2025alphaevolve} attack open problems through specialized or evolutionary search, though their outputs are constructions and programs, not machine-checked proofs. Benchmarks confirm the distance to research difficulty~\citep{glazer2024frontiermath}. General models now contribute steps to live mathematics: a Gemini system surveyed hundreds of Erd\H{o}s problems~\citep{aletheia} and GPT-5 produced new results with mathematicians~\citep{bubeck2025gpt5science}. Both episodes drew scrutiny over whether claims were proved or merely retrieved, which is exactly what a formal proof settles~\citep{tao2025machineassisted,bloom}. 
AlphaProof Nexus~\citep{tsoukalas2026formalproofsearch} is an evolutionary Lean proof-search system: prover subagents edit marked regions of Lean sketches, invoke AlphaProof on subgoals, validate candidates, and use rater agents with a population database to select promising sketches, resolving $9$ of $353$ Erd\H{o}s problems and $44$ of $492$ OEIS conjectures. In contrast, \agentname targets paper-level autoformalization: it builds an audited blueprint DAG from source and proves the resulting multi-result Lean development bottom-up.

\paragraph{Autonomous formalization agents and large-scale formalization.}
A wave of 2025 and 2026 systems formalize at the scale of whole results or papers, the regime \agentname targets. 
AxiomProver~\citep{axiommath2025}, Gauss~\citep{hariharan2026gauss}, and Aristotle~\citep{aristotle} are capable to formalize papers but they are fully close-source company products. However, we have no clues about such harness design.
Besides, large-scale community projects are growing: the Polynomial Freiman-Ruzsa conjecture~\citep{pfr2023}, the Liquid Tensor Experiment~\citep{scholze2022liquid}, Fermat's Last Theorem~\citep{buzzard2024flt}, Carleson's theorem~\citep{becker2024carleson}, the Equational Theories Project~\citep{bolan2025equational}, statistical learning theory~\citep{sonoda2025lean,zhang2026statistical}, and reinforcement learning theory~\citep{zhang2025towards}.

We present \agentname in two parts: its infrastructure design in \cref{sec:harness} which introduces the blueprint and the agents that act on it, and then the orchestration that drives them to be a multi-hour, parallel autoformalization recoverable and resistant to drift in \cref{sec:method-orchestrator}. \cref{sec:experiments} provides the experimental evaluations. The conclusion is drawn in \cref{sec:conclusion}.
\section{Harness Infrastructure}
\label{sec:harness}

This section covers our harness's fundamental ingredients. \cref{sec:method-substrate} describes the blueprint format we use. \cref{sec:method-agents} then introduces the four contract-scoped agents that maintain the blueprint, each owning a well-separated task, skilled workflow, and bounded edit scope.

\subsection{The Blueprint as the System of Record}
\label{sec:method-substrate}

The central artifact in our harness is the blueprint, a Lean file, that serves simultaneously as a formal proof skeleton, a natural-language proof graph, and the task interface exposed to agents. All durable mathematical state is stored in this file: theorem statements, intermediate lemmas, definitions, prose explanations, and declared proof dependencies. Agents may read, prove, review, or repair parts of the blueprint, but they do not maintain any hidden shared memory outside it.

Each mathematical node in the blueprint is represented by a Lean declaration annotated with structured metadata, in the format of LeanArchitect~\citep{zhu2026leanarchitect}: the blueprint metadata lives in an in-source \code{@[blueprint ...]} attribute, and the dependency graph and \code{sorry}-status are inferred from Lean's elaborator. This complements the original blueprint methodology~\citep{massot2020leanblueprint}. A typical proof node has the form:
\begin{center}
\begin{minipage}{0.95\textwidth}
\begin{lstlisting}[style=lean]
@[blueprint "lem:weighted-tail-bound"
  (statement := /-- LaTeX statement text -/)
  (proof     := /-- LaTeX proof prose with \cref{...} citations    -/)
  (title     := /-- one-line LaTeX title                           -/)
  (latexEnv  := "lemma")]
lemma weighted_tail_bound ... : ... := by
  sorry_using [aux_lemma_one, aux_lemma_two]
\end{lstlisting}
\end{minipage}
\end{center}
The Lean declaration gives the formal statement that must type-check. The \code{statement} field records the corresponding mathematical statement in LaTeX. The \code{proof} field records the natural-language proof sketch, including explicit references to earlier nodes. The \code{title} and \code{latexEnv} fields provide presentation metadata and allow the verifier to check that the Lean declaration and the prose environment agree.

Proof nodes may have one of three proof-body states: unproved, unproved but with a dependency list, and proved. A node may be unproved, written as \code{by sorry}; it may be unproved but equipped with an intended dependency list, written as \code{by sorry\_using [...]}; or it may contain a complete Lean proof. Definitions are treated as global context and are not proof nodes in the dependency DAG. The proof DAG is formed only from lemma and theorem declarations.
Note that, \code{\textbackslash cref\{...\}} citations in the LaTeX prose are not decorative: the verifier extracts the actual proof dependencies via Lean's elaborator metadata and enforces \emph{two-way} parity between the \code{\textbackslash cref} edges and the elaborator's edges. The blueprint obtains the property that the natural-language graph and the typed graph cannot drift apart: a PR that lets either side disagree with the other is rejected before it merges.

\subsection{Contract-Scoped Agents}
\label{sec:method-agents}

Our harness decomposes research-level autoformalization into four contract-scoped agents: the \agent{Blueprinter}, the \agent{\agent{Target-Reviewer}}, the \agent{Worker}, and the \agent{Refiner}. Each agent owns a narrow interface, a bounded edit scope, and a specific failure mode that it is designed to expose or contain. 
\cref{tab:agents} summarizes the four agents along these axes.

\begin{table}[t]
\centering
\small
\setlength{\tabcolsep}{5pt}
\caption{The four contract-scoped agents. Only the \agent{Blueprinter} and \agent{Refiner} are given the source proof; the \agent{Target-Reviewer} and \agent{Worker} see only the canonical statements and the blueprint. Every PR reaches \code{main} only through the CI verifier.}
\label{tab:agents}
\begin{tabular}{@{}l p{3.0cm} l p{3.0cm} p{3.5cm}@{}}
\toprule
\textbf{Agent} & \textbf{Input} & \textbf{Output} & \textbf{Allowed edits} & \textbf{Failure mode} \\
\midrule
\agent{Blueprinter} & source proof, canonical statements, blueprint & PR & writes the whole skeleton, bodies as placeholders & poor decomposition, i.e.\ a large repair radius \\
\addlinespace
\agent{Target-Reviewer} & canonical statements, blueprint & 
issue/None & none (read-only) & a misformalized target: a valid but wrong theorem \\
\addlinespace
\agent{Worker} & canonical statements, blueprint & PR/issue & the node's prose, its proof body, its local refinement region & a local proof failure, or silently proving a misformalized node \\
\addlinespace
\agent{Refiner} & source proof, canonical statements, open issues, blueprint & PR & one connected illness sub-DAG & blueprint drift and source gaps \\
\bottomrule
\end{tabular}
\end{table}

This separation is essential: a single agent asked to design the proof graph, judge its own fidelity, discharge proofs, and repair global errors has no reliable mechanism for detecting its own drift. In our harness, every agent produces an externally checkable artifact, and every artifact is accepted only through the verifier.

\subsubsection{\agent{Blueprinter}}

The \agent{Blueprinter} converts the source proof and the canonical target statements into the initial blueprint. Its job is not to prove the paper but to choose a decomposition of the argument that is faithful enough for later review and local enough for later repair. A good blueprint should isolate mathematical commitments so that, if one statement is later found to be misformalized, the number of downstream declarations that must be changed is small.

We frame the \agent{Blueprinter}'s job as a \emph{repair-radius optimization problem}: draft an initial blueprint which minimizes the expected number of declarations that must change if one declaration turns out to be wrong. The decomposition follows a published rubric via \code{decomposition.md}.
The \agent{Blueprinter} writes every proof body as \code{sorry} or \code{sorry\_using}, and delivers a single PR to merge the blueprint into the \code{main} branch. The agent is supposed to exit after PR creation, then the \textbf{stop hook} is triggered to monitor the merge status. If merge fails, the stop hook will block the exit, serve a resume instruction with CI run's job logs.
Mathematical repair of the source is explicitly out of scope, the agent's job is to \emph{isolate} it, not to \emph{fix} it. This boundary design provides a clear separation with later \agent{Refiner} and allocate more agent's workloads for decomposition. The in-workspace knowledge-store layout of \agent{Blueprinter} is given in \cref{tree:blueprinter} from Appendix~\ref{app:agent-layouts}.

\subsubsection{\agent{Target-Reviewer}}

The \agent{\agent{Target-Reviewer}} audits the blueprint before large-scale proving begins. Its role is to prevent the most expensive failure mode in research-level autoformalization: proving a formally valid Lean theorem that is not the theorem intended by the source paper.
For instance, if a root \code{theorem} statement is misformalized, every \code{lemma} the \agent{Worker} proves afterwards is wasted compute, so an early certification at the roots is what makes later orchestration loop more productive.

For every target theorem, the \agent{Reviewer} compares three objects: the canonical target statement, the LaTeX statement stored in the blueprint, and the Lean type. It checks whether they express the same mathematical claim, with the same hypotheses, quantifiers, definitions, and conclusion.

The \agent{Reviewer} is not allowed to edit the blueprint. A clean review allows our harness to enter the proof-discharge stage. A failed review produces a grouped issue describing the suspected mismatch and the affected nodes; the issue is then handled by the \agent{Refiner}.

\subsubsection{Per-node \agent{Worker}}

\begin{wrapfigure}[16]{r}{0.35\textwidth}
  \centering
  \vspace{-0.5cm}
  \includegraphics[width=\linewidth]{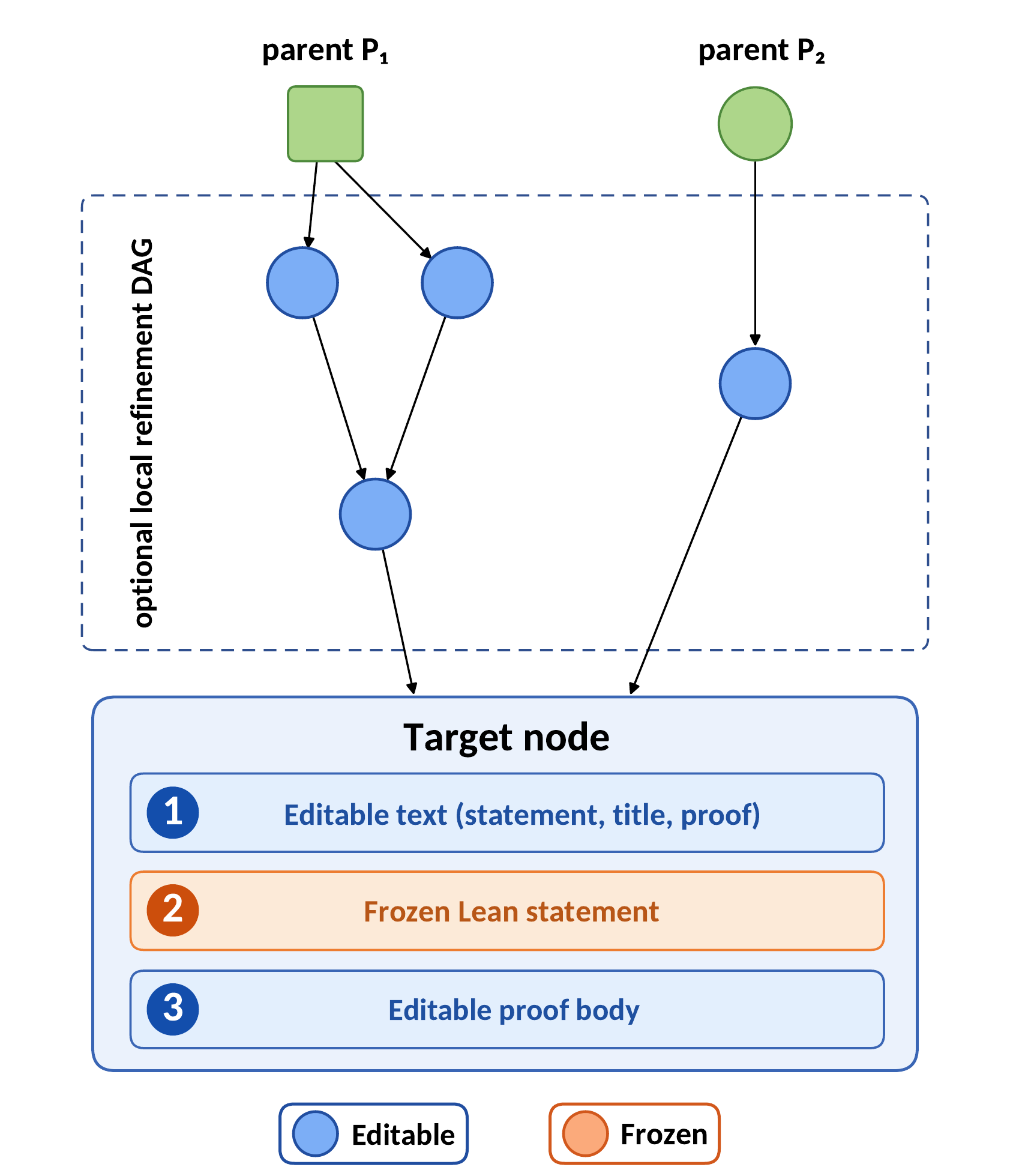}
  \caption{Expected editing behaviour.}
  \label{fig:exp-edit}
\end{wrapfigure}
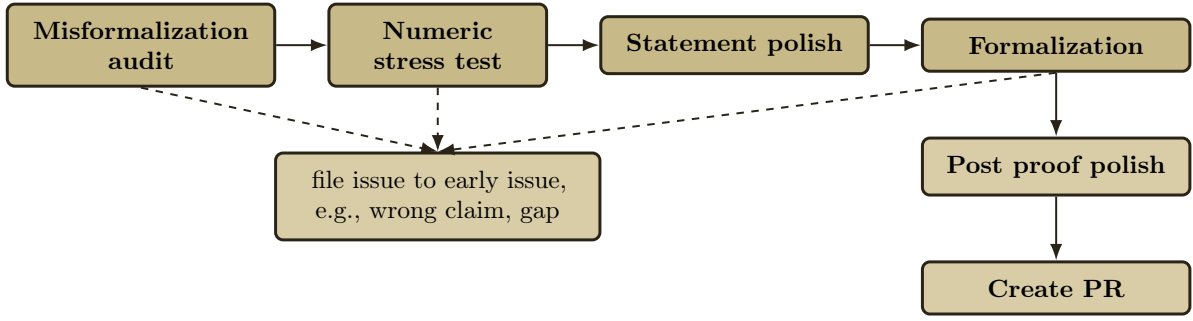
\begin{figure}[t]
  \centering
  \resizebox{0.95\linewidth}{!}{%
  \begin{tikzpicture}[node distance=0.72cm]
    \node[gate,text width=3.2cm] (p1) {Misformalization audit};
    \node[gate,right=of p1,text width=2.5cm] (p2) {Numeric stress test};
    \node[gate,right=of p2,text width=3.2cm] (p3) {Statement polish};
    \node[gate,right=of p3,text width=3.2cm] (p4) {Formalization};
    \draw[arrow] (p1) -- (p2);
    \draw[arrow] (p2) -- (p3);
    \draw[arrow] (p3) -- (p4);
    \node[flow,below=0.9cm of p2,text width=4cm,fill=TanLt] (issue) {file issue to early issue,\\e.g., wrong claim, gap};
    \node[gate,below=0.9cm of p4,text width=3.2cm,fill=TanLt] (post) {Post proof polish};
    \draw[arrow] (p4) -- (post);
    \node[gate,below=0.9cm of post,text width=3.2cm,fill=TanLt] (pr) {Create PR};
    \draw[arrow] (post) -- (pr);
    \draw[dashedarrow] (p1.south) -- (issue.north);
    \draw[dashedarrow] (p2.south) -- (issue.north);
    \draw[dashedarrow] (p4.south) -- (issue.north);
  \end{tikzpicture}}
  \caption{The executable workflow of per-node \agent{Worker}.}
  \label{fig:worker-workflow}
\end{figure}

A \agent{Worker} is assigned one proof node whose declared dependencies have already been proved. Its job is local: either prove the assigned Lean statement or report why the node should not yet be proved. The \agent{Worker} proceeds through four ordered phases, see \cref{fig:worker-workflow}.

\paragraph{Phase 1: Misformalization audit.} The \agent{Worker} first treats the Lean type as a suspect specification. It compares the Lean statement with the blueprint prose, checks why the node exists, identifies which downstream nodes use it, and verifies that the statement provides the fact those downstream nodes need. If the type is missing a hypothesis, states the wrong conclusion, uses an unsuitable abstraction, or does not match its intended role in the proof DAG, the \agent{Worker} stops and files an issue.

\paragraph{Phase 2: Cheap falsification.} When the claim admits finite, numerical, or boundary-case testing, the \agent{Worker} attempts to refute it before spending prover compute. Failure to find a counterexample is not treated as a proof; it is only a low-cost sanity check. A discovered counterexample or suspicious boundary case is reported as an issue.

\paragraph{Phase 3: Statement polish.} If the Lean type passes the first two phases, the \agent{Worker} may edit only the node’s prose fields: the LaTeX statement, title, and proof text. The goal is to make the natural-language text describe the Lean statement exactly, neither stronger nor weaker.

\paragraph{Phase 4: Formalization.} The \agent{Worker} then attempts to replace the placeholder body with a complete Lean proof. The Lean type of the assigned node is frozen throughout this phase. The \agent{Worker} may introduce fresh helper lemmas only inside the local refinement region immediately before the target node. These helpers must precede the target, depend only on already visible declarations or earlier local helpers, and terminate at the assigned target. If \agent{Worker} cannot complete the formalization within all boundaries, then it needs to file an issue with clear evidence to exit.

During formalization, The \agent{Worker}’s edit scope is mechanically enforced via an editing MCP server built by patching Codex's \code{apply-patch} tool, illustrated in \cref{fig:exp-edit}. It may edit the assigned node’s prose fields, its proof body, and its local refinement region, but not the Lean type of the target or unrelated blueprint nodes. This restriction makes parallel formalization safe: different \agent{Worker}s operate on disjoint editable regions, so successful patches commute by construction, while failed attempts become rejected local patches or diagnostic issues rather than global corruption.
To be specific, the Lean file is partitioned around the assigned target node $T$ into frozen and editable spans:
\begin{center}
\begin{minipage}{0.95\textwidth}
\begin{lstlisting}[style=lean]
-- previous node ends here
-- BEGIN editable local refinement area for T
@[blueprint "lem:X" ...]
lemma X ... := by
...

@[blueprint "lem:Y" ...]
lemma Y ... := by
...
-- END editable local refinement area for T
@[blueprint "lem:T" -- frozen
  (statement := /-- editable -/)
  (proof := /-- editable -/)
  (title := /-- editable -/)
  (latexEnv := "lemma")] -- frozen
lemma T ... : ... := by -- frozen
-- editable proof body
\end{lstlisting}
\end{minipage}
\end{center}

This harness design makes the parallel multi-agent loop against the same frozen substrate commit works. \agent{Workers} acting on disjoint editable regions produce patches that commute by construction so PRs can land in any order without merge conflict. Within its editable region, the \agent{Worker} may \emph{grow a local refinement DAG} consisting of fresh helper nodes before the target node which is the unique terminal, which aims to overcome the possible risk of under-decomposition brought from \agent{Blueprinter} or \agent{Refiner}.

\subsubsection{\agent{Refiner}}
\label{sec:refiner}

The \agent{Refiner} repairs blueprint-level defects reported by the \agent{Target-Reviewer} or \agent{Worker}s. Unlike a \agent{Worker}, which is restricted to one proof node, the \agent{Refiner} can edit the whole blueprint since a defect might affect multiple declarations.

Given the open issue(s), the \agent{Refiner} first identifies the affected region, called the {\bf illness} area: the smallest connected sub-DAG that must be inspected or changed to resolve the defect. We develop an MCP server called \code{dag-tracker} for agent to call for live parent/child identification.

The \agent{Refiner} classifies each defect as either {\bf blueprint drift} or a {\bf source gap}. Blueprint drift means that the Lean blueprint has diverged from the source proof: for example, a statement was misformalized, a dependency was wrong, or the prose no longer describes the Lean declaration. A source gap means that the source proof itself is incomplete, ambiguous, or false at the level required for formalization.
During the repair phase, the proof bodies are governed by the following decision tree node by node:
\vspace{0.2cm}
\dirtree{%
.1 Is the node NEW or EXISTING?.
.2 NEW: body as placeholder.
.2 EXISTING: what is the current proof body's shape?.
.3 placeholder: keep and align with new dependency set if prose changed..
.3 complete tactic proof: still compiles after repair?.
.4 YES: preserve the complete proof body byte-identical to the input blueprint..
.4 NO: wholesale-replace with placeholder.
}
\vspace{0.2cm}
\noindent Compilation is decided by Lean compiler, never by the agent. Wholesale replacement is recorded in the PR summary and never negotiated. There is no path that lets the \agent{Refiner} partial-edit a complete proof body. Although a downstream \agent{Worker} needs to prove the downgraded node again, the discipline keeps the \agent{Refiner}'s blast radius proportional to the issues it is closing, not to the file it is editing. Furthermore, the source gap should be fixed via reasoning and the new solution should be updated in the LaTeX fields.

\section{System Orchestration}
\label{sec:method-orchestrator}

After introducing harness infrastructure, we are ready to discuss our system orchestration in two stages. The first stage constructs and audits a faithful blueprint before formalization begins. The second stage discharges the proof DAG through parallel, CI-gated proof attempts.

\subsection{Stage~1 -- Cold Start and Target Review} 

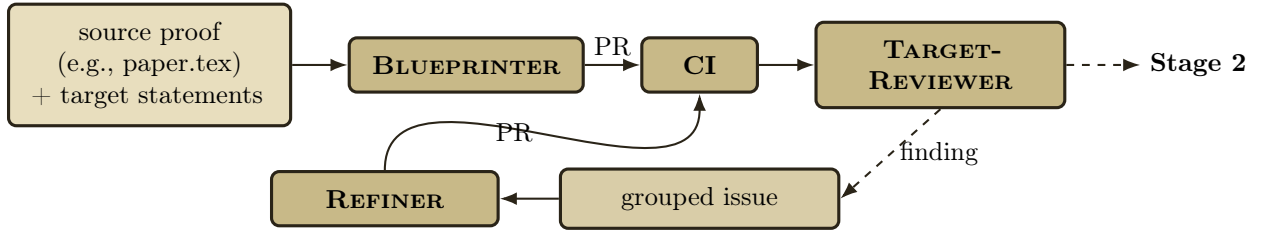
\begin{figure}
  \centering
  \resizebox{\linewidth}{!}{%
   \begin{tikzpicture}[node distance=0.75cm]
    \node[flow,text width=3.25cm] (source) {source proof (e.g., paper.tex)\\+ target statements};
    \node[gate,right=of source,text width=2.6cm] (bp) {\agent{Blueprinter}};
    \node[gate,right=of bp,text width=1.0cm] (ci) {CI};
    \node[gate,right=of ci,text width=2.8cm] (review) {\agent{Target-Reviewer}};
    \node[flow,below=1.0cm of ci,text width=3.2cm,fill=TanLt] (issue) {grouped issue};
    \node[gate,left=0.8cm of issue,text width=2.5cm] (ref) {\agent{Refiner}};
    \draw[arrow] (source) -- (bp);
    \draw[arrow] (bp) -- node[above]{PR} (ci);
    \draw[arrow] (ci) -- (review);
    \draw[dashedarrow] (review.south) -- node[right]{finding} (issue.east);
    \draw[arrow] (issue) -- (ref);
    \draw[arrow] (ref.north) to[out=90,in=-90] node[left]{PR} (ci.south);
    \draw[dashedarrow] (review.east) -- ++(1.0,0) node[right]{\textbf{Stage 2}};
   \end{tikzpicture}}
  \caption{The orchestration of stage 1.}
  \label{fig:stage-1}
\end{figure}

We frame this stage as a \textbf{nested Ralph-Wiggum loop}, shown in \cref{fig:stage-1}, which is designed to digest the input natural language source into an initial Lean blueprint. 

The \agent{Blueprinter} produces an initial Lean blueprint in which all declarations elaborate and all proof bodies are placeholders which must pass the CI gate. The \agent{Target-Reviewer} then compares the theorem nodes against the canonical targets and the blueprint's own LaTeX and Lean.

If the review is clean, our harness proceeds to Stage 2. If the Reviewer finds a target mismatch, missing hypothesis, incorrect dependency, or other blueprint-level defect, it files a grouped issue. The \agent{Refiner} repairs the affected region and submits a repair PR. After the PR passes CI and merges, the Reviewer runs again. Stage 1 terminates only when the target review exits clean.

\subsection{Stage~2 -- DAG-orchestrated Loop}

\begin{figure}[ht]
    \centering
    \includegraphics[width=\linewidth]{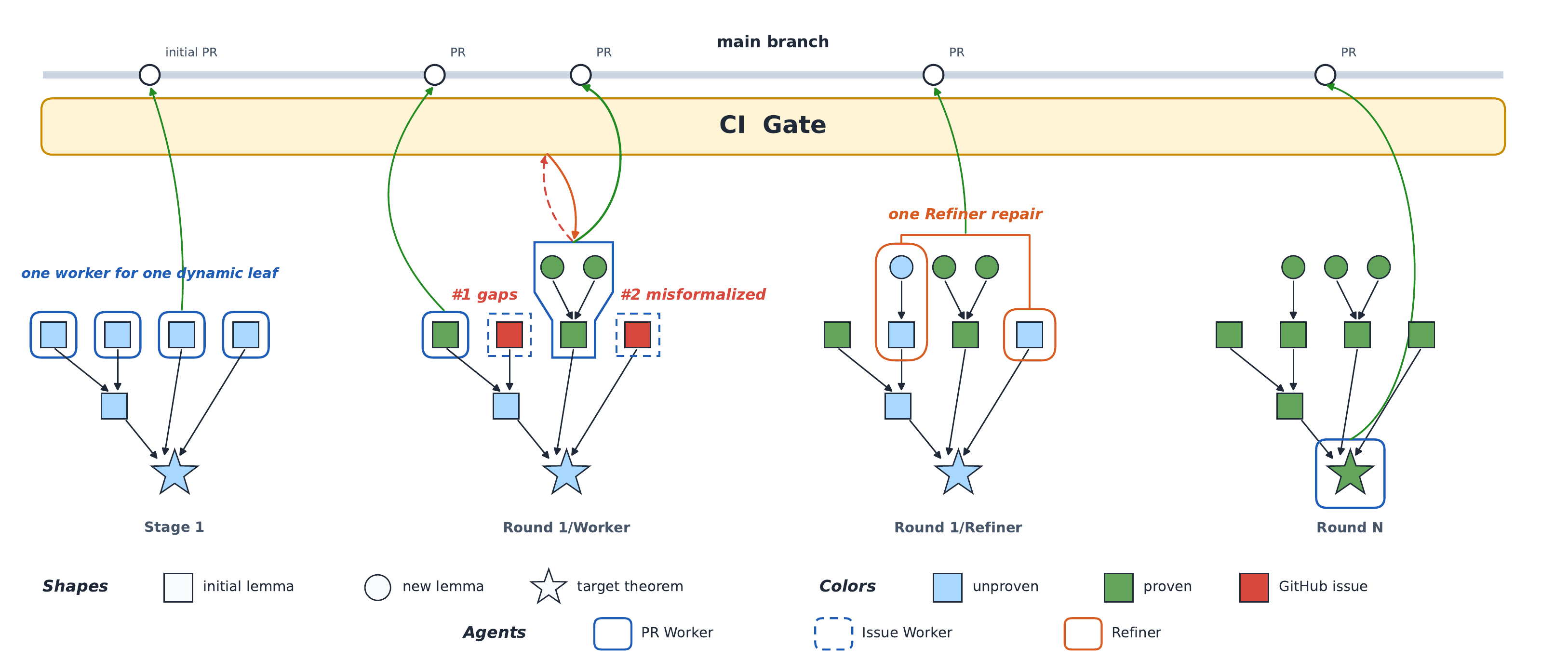}
    \caption{The DAG-based orchestration of stage 2.}
    \label{fig:stage-2}
\end{figure}

Stage 2 proves the blueprint via the pipeline in \cref{fig:stage-2}. Each round starts from the current \code{main} branch, which is the system of record. The orchestrator extracts the proof DAG from the blueprint and identifies the current dynamic leaves: unproved proof nodes whose dependencies have already been proved.

The orchestrator assigns each dynamic leaf to an independent \agent{Worker}. Every \agent{Worker} receives the same frozen substrate commit and a mechanically restricted editable region around its target node. \agent{Worker}s run in parallel. A successful \agent{Worker} submits a proof PR; otherwise it files an issue instead.

Proof PRs are accepted only by the CI gate.Passing PRs are merged into main; failing PRs are rejected with diagnostics. After all \agent{Worker}s in a round have finished, the \agent{Refiner} processes the accumulated issues, submits repair PR through the same CI gate, and updates the blueprint if the repairs pass.

Our harness repeats these rounds until every proof node in the DAG has a complete Lean proof. At termination, the main branch contains a fully proven blueprint with no remaining placeholders.

\paragraph{Why direct merge, not Github auto-merge.} The orchestrator merges via \code{gh pr merge --squash} directly, not \code{gh pr merge --auto --squash}. Auto-merge requires a pending required check (branch protection) and serialises across parallel PRs, i.e., only the first PR to finish CI can enable auto-merge, and a later parallel PR finds \code{main} has moved and the GraphQL mutation refuses. Direct merge needs no pending check and lets every non-conflicting parallel PR land independently as its own CI completes.

\subsection{Sustaining hours-long runs}
\label{sec:method-long-running}

The following composable properties keep a multi-hour run resumable without drift.

\paragraph{CI verifier as continuous integration}
The CI verifier is the single gate for any merge request to \code{main} branch.
The verifier encodes the blueprint contract (\code{blueprint-format.md}) as the following seven executable checks:
\begin{enumerate}
\item \textbf{Lean compilation}: every diagnostic must be either none, or \texttt{declaration uses 'sorry'} warning. 
\item \textbf{Node well-formedness}: every \code{@[blueprint]} attribute has non-empty \code{statement} / \code{title} / \code{proof} fields. Placeholders are multi-line \code{by} followed by \code{sorry} or \code{sorry\_using}. Incomplete proof body is prohibited. 
\item \textbf{\code{latexEnv} consistency:} the Lean keyword and the \code{latexEnv} field must agree (e.g. \code{lemma}~$\leftrightarrow$~\code{lemma}, \code{theorem}~$\leftrightarrow$~\code{theorem}).
\item \textbf{Label-name normalization}: blueprint labels like \code{lem:foo-bar} normalize (\code{-}~$\to$~\code{\_}) to the actual Lean name. 
\item \textbf{Unique labels}: each blueprint node's Lean naming should be unique.
\item \textbf{Proof-dependency parity}: Based on the proof DAG exatracted via Lean's elaborator metadata, a \emph{two-way} parity then requires every Lean dep to be \code{\textbackslash cref}-cited in the prose and every \code{\textbackslash cref\{lem:\_\}} or \code{\textbackslash cref\{thm:\_\}} to be a Lean dependency. \code{sorry\_using} references additionally must precede the citing node in file order and must point at proof nodes only. We treat the definitional nodes as global context which are intentionally excluded from the proof DAG.
\item \textbf{Lemma closeness}: every \code{lemma} must be cited by some later \code{lemma} or \code{theorem}; a \code{theorem} is a target claim and is treated as a terminal sink, allowed to have no children. Equivalently, every non-terminal node of the proof DAG must have positive out-degree toward a target, so no \code{lemma} is an \emph{orphan} that feeds nothing. Inspired by the \emph{logical closeness} defined by~\citet{zhang2026dagmath}, which requires every non-sink node of a reasoning DAG to have positive out-degree, this check is a \emph{structural} guard against task drift: an orphan lemma is exactly the footprint a drifting agent leaves when it proves machinery absent from the source, the failure mode that stalled our earlier harness (\cref{sec:exp-ablation}).
\end{enumerate}
Either successful or failed verification triggers an upserted PR comment for live feedbacks. These checks are deliberately structural rather than stylistic/semantic. Lemma closeness in particular is the structural counterpart to feeding the \agent{Refiner} the source proof (\cref{sec:refiner}): source-anchored repair realigns proof \emph{content} to the paper, whereas lemma closeness enforces goal-directedness on the dependency \emph{graph} alone, mechanically and without reading the mathematics. 

This harness does not try to prescribe the exact tactic script a \agent{Worker} should discover; instead, it enforces the interfaces that make independent agents composable. Formal statements must compile, prose citations must match proof dependencies, graph edges must be visible and acyclic, and every pull request must preserve the blueprint as a coherent system of record. This is the main methodological advantage of our harness: it converts long-running agent work from an opaque sequence of prompts into a set of small, recoverable, mechanically checked transactions.

\paragraph{Stop-hook-driven self-recovery.} Each PR-enabled agent's stop hook runs after the agent declares itself done. The hook validates \code{delivery.yml} as the terminal-delivery record; if the agent opened a PR, the hook polls GitHubfor either a successful merge or a CI failure. On a CI failure the hook extracts the verifier's upserted comment and the raw failed-job logs as context inside the worktree, and blocks the agent with a templated debug-fix-push instruction.

\paragraph{Bounded scope, deterministic enforcement.}
Every agent receives a contract that names the spans of the workspace it may edit. The contracts are not advisory: the \code{apply-patch} MCP server rejects every patch to a frozen span, the \code{read-only} Codex sandbox rejects every constructive operations, and the CI path-allowlist rejects every PR that touches a path outside blueprint. The principle averts the failure mode in which one agent's mistake silently corrupts another agent's working state. The worst outcome any misbehaving agent can produce is a rejected patch, never a poisoned PR.

\paragraph{Context management.}
All inter-agent communication flows through the on-disk Lean blueprint and the GitHub PR/issue stream. There is no scratch memory, no shared in-RAM channel, no out-of-tree file written by one agent and read by another. The principle averts \emph{cross-agent context bleed}: with no hidden channel, an agent that misbehaves cannot poison another agent's context, and a Codex auto-compaction that wipes one agent's window does not corrupt anything any other agent will run on.
\section{Experiments}
\label{sec:experiments}

We test \agentname on two research papers \citep{erdosGraham2026,primitiveChains2026} on Erd\H{o}s problems. Each input is the paper's LaTeX source together with a separate file of canonical target statements; each output is a Lean blueprint in which every proof node is proven (\cref{sec:harness}). The harness formalized all seven target theorems across the two papers, covering four Erd\H{o}s problems (\#1051, \#1196, \#164, \#1217), with no \code{sorry} and under the seven-check CI contract of \cref{sec:method-long-running}. It discharged 258 lemmas and theorems in total. Aristotle~\citep{aristotle}, the only commercial Lean agent we could access, failed on both papers after tens of hours.

\subsection{Two research papers for evaluation}
\label{sec:exp-benchmark}

We chose two papers that are recent, research-level, and AI-assisted in their genesis. Difficulty rules out competition-style targets: both papers are analytic number theory with multi-page estimates, not some tricks. AI provenance makes formalization the decisive artifact: both proofs were \emph{discovered} with AI help, and an AI-discovered proof is exactly the case where a machine-checked formalization, rather than prose, is the guarantee that the argument is correct.\footnote{In the broader Erd\H{o}s-problem sweep that produced one of our targets, DeepMind reported that of 200 candidate solutions marked correct or incorrect, only $6.5\%$ were ``meaningfully correct''~\citep{aletheia}. This gap is what a \code{sorry}-free Lean proof closes.}

\paragraph{Erd\H{o}s--Graham irrationality~\citep{erdosGraham2026}.} This paper answers Erd\H{o}s Problem \#1051, posed by Erd\H{o}s and Graham in 1980~\citep{bloom}: a double-exponential growth condition forces $\sum_n 1/(a_n a_{n+1})$ to be irrational. The headline case was solved autonomously by the agent \emph{Aletheia} (Gemini Deep Think), one of four Erd\H{o}s problems DeepMind reported solving autonomously in December 2025~\citep{aletheia}. We stress that \#1051 is the \emph{easy} part. Its $d=2$ golden-ratio instance follows from a short Borel-peak argument; the depth of the paper lies in two later results that we also formalize. The \emph{general} irrationality theorem fixes a weight tuple $\mathbf{w}$, defines a growth threshold $c_{\mathbf{w}}$ as the unique positive root of a polynomial $P_{\mathbf{w}}$, and proves irrationality of a weighted series via Mahler's criterion, a local-peak selection lemma, and a three-regime tail analysis. The \emph{construction} theorem is its sharp negative counterpart: for every $C>1$ it builds an integer sequence whose weighted sum is rational, through a non-constructive nested-interval covering argument. The original question already drew formalization attention, but these two generalizations were unformalized.

\paragraph{Primitive sets and von Mangoldt chains~\citep{primitiveChains2026}.} This paper introduces a method, suggested by output of GPT-5.4 Pro, that bounds Erd\H{o}s sums of primitive sets using Markov chains with von Mangoldt weights on the divisibility poset. The single method resolves several conjectures at once. We target three. Erd\H{o}s--S\'ark\"ozy--Szemer\'edi \#1196 (a 1966 conjecture) asserts $f(A)\le 1+O(1/\log x)$ for a primitive set $A\subseteq[x,\infty)$, sharpening the previous record of $\approx 1.399$ to the conjectured constant $1$. The Erd\H{o}s Primitive Set Conjecture \#164 asserts $f(A)\le f(\mathbb{N}_1)=1.6366\dots$. Erd\H{o}s--S\'ark\"ozy--Szemer\'edi \#1217 produces an infinite divisibility chain inside any set of positive doubly-logarithmic density. The proofs use the sub-invariance of the doubly-harmonic weight under the von Mangoldt chain, Mertens-type estimates, and the monotonicity of the Dirichlet eta function. This paper is a strong reference point because parts of it were already formalized by others: \#1196 by Math Inc.'s \emph{Gauss} agent in roughly $4{,}000$ lines of Lean~\citep{gauss1196}, and \#164 by Boris Alexeev using Codex~\citep{primitiveChains2026}. Conjecture \#1217 had not been formalized by anyone.

\subsection{Setup}
\label{sec:exp-setup}

We ran the harness three times; every agent runs on Codex (GPT-5.5-xhigh, 258K, read-only, no web access). For Erd\H{o}s--Graham, one run (\code{ErdosGraham}\footnote{\url{https://github.com/YuanheZ/ErdosGraham}}) formalizes all four target theorems. For the primitive-sets paper we split the work into two runs to test \emph{incremental development}: whether a finished formalization can be extended by adding targets to the problem file and rerunning the harness, rather than starting over. A first run (\code{Erdos1196}\footnote{\url{https://github.com/YuanheZ/Erdos1196}}) formalizes \#1196. We then seed a second repository (\code{Prim}\footnote{\url{https://github.com/YuanheZ/Prim}}) with the final \#1196 blueprint, add \#164 and \#1217 to the target statements, and rerun the harness; both build on the \#1196 infrastructure. Each run receives only the paper source and the target statements; success means the \code{main} branch reaches a state where every blueprint proof node carries a complete proof, no \code{sorry} or \code{sorry\_using} remains, and CI passes all seven structural checks (\cref{sec:method-long-running}).

\subsection{Results}
\label{sec:exp-results}

\paragraph{Every target is formalized.} \agentname discharged all seven target theorems with complete, machine-checked proofs (\cref{tab:outcome}): every run is \code{sorry}-free, with no \code{axiom} or \code{native\_decide}.

\begin{table}[t]
  \centering
  \caption{Formalization outcomes. Proof nodes counts \code{lemma} and \code{theorem} declarations; definitions are global context. The \code{Prim} row includes the $59$-node \#1196 blueprint reused as its seed.}
  \label{tab:outcome}
  \begin{tabular}{l|ccc}
    \toprule
    & \textbf{Erd\H{o}s--Graham} & \textbf{\#1196} & \textbf{\#164 \& \#1217} \\
    \textbf{Metric} & (\code{ErdosGraham}) & (\code{Erdos1196}) & (\code{Prim}) \\
    \midrule
    Target theorems            & 4 & 1 & 2 \\
    Lean lines                 & 8{,}513 & 3{,}988 & 14{,}592 \\
    Nodes (def\,/\,lem\,/\,thm) & 39\,/\,106\,/\,5 & 15\,/\,43\,/\,1 & 57\,/\,144\,/\,3 \\
    Proof nodes                & 111 & 44 & 147 \\
    Remaining \code{sorry}     & 0 & 0 & 0 \\
    Status                     & \textbf{complete} & \textbf{complete} & \textbf{complete} \\
    \bottomrule
  \end{tabular}
\end{table}

\paragraph{Incremental development works.} Seeding \code{Prim} with the final \#1196 blueprint, the harness reused all $59$ nodes unchanged and added $145$ new nodes ($103$ new proof obligations) to discharge \#164 and \#1217, reaching $14{,}592$ lines over $204$ nodes. It extended a finished formalization to new targets rather than restarting, confirming the incremental-development mode \cref{sec:exp-setup} set out to test. Across the three runs the harness proved $258$ distinct lemmas and theorems: $111$ for Erd\H{o}s--Graham, $44$ for \#1196, and $103$ more for \#164 and \#1217.

\paragraph{Formalization feedback sharpens the mathematics.} The Lean compiler is not only a checker but also can provide ground-truth mathematical signal, and the issue-to-\agent{Refiner} loop turns that signal into better mathematics. Across the three runs essentially every blocked-node issue is grounded in a concrete formalization artifact rather than a prose disagreement, and three kinds recur. The compiler \emph{refutes false statements}: a \agent{Worker} instantiates a tail estimate and collapses its conclusion to $1\le 0$ (\href{https://github.com/YuanheZ/ErdosGraham/issues/431}{issue~\#431}), or a two-window inequality to $4/3\le 2/3$ (\href{https://github.com/YuanheZ/ErdosGraham/issues/465}{issue~\#465}), each forcing a corrected hypothesis. Mathlib's totalization conventions \emph{expose vacuous targets} that prose hides: a non-summable real \code{tsum} is defined to be $0$, so the formalized \#1196 bound held even for a divergent series until a \code{Summable} hypothesis was added (\href{https://github.com/YuanheZ/Prim/issues/2}{issue~\#2}), and a real \code{limsup} of an unbounded count returns $0$, so the divisibility-chain density target was unsound until it was recast in \code{ENNReal} (\href{https://github.com/YuanheZ/Prim/issues/102}{issue~\#102}). And failed tactic probes together with Mathlib-absence findings \emph{locate the missing mathematics}: when \code{positivity} cannot sign the derivative of the Dirichlet eta function and no monotonicity lemma exists (\href{https://github.com/YuanheZ/Erdos1196/issues/16}{issue~\#16}), and the naive route then reduces to a false inequality between Gamma measures (\href{https://github.com/YuanheZ/Erdos1196/issues/20}{issue~\#20}), the \agent{Refiner} is driven to the paper's actual argument, eta monotonicity via stochastic domination of Gamma laws. A human reads ``$f(A)\le\cdots$'' charitably, assuming the series converges; the formalization does not, and that mechanized skepticism, surfaced as an issue and resolved by the \agent{Refiner}, is where much of the harness's mathematical value lies.

\paragraph{The runs are autonomous.} \cref{tab:orchestration} reports the orchestration cost of each fully autonomous run. The Erd\H{o}s--Graham run took $19$ rounds, launched $58$ \agent{Worker}s and $7$ \agent{Refiner}s, and cost \$257 in GPT-5.5 API-equivalent tokens; the \#1196 run took $17$ rounds and \$189; the \#164/\#1217 run was the largest, at \$624. The CI gate is selective rather than ceremonial: of the $58$ Erd\H{o}s--Graham \agent{Worker}s, $44$ produced PRs that passed all checks and merged, the rest being rejected before reaching \code{main}.

\paragraph{Parallel PRs never conflict.} Across the three runs, $135$ \agent{Worker} pull requests landed on a continuously moving \code{main} by direct squash-merge, as many as $16$ in a single round's parallel batch, and not one produced a merge conflict. This realizes the central guarantee of the frozen editable region (\cref{sec:method-agents}): \agent{Worker}s acting on disjoint spans produce patches that commute, so PRs land in any order without collision.

\paragraph{Human observation tightens the harness.} We observe the agent's working trace to constantly reshape the design. Early Erd\H{o}s--Graham runs fixed the \code{maxHeartbeats} to $0$, which disables Lean's deterministic timeout. Reading the \agent{Worker} traces, we found a strong preference for automation (\code{nlinarith}, \code{simp}, \code{aesop}) over explicit tactic scripts. With no deterministic ceiling, such a search runs until a coarse external timeout cuts the whole build, stalling a node with no reproducible signal. We changed the header to a fixed $500$K-budget, restoring a per-declaration deterministic ceiling: a runaway search now fails fast and locally instead of exhausting wall-clock time, and the agent's reliance on unbounded automation dropped sharply. To make efficient proofs the path of least resistance, we gave the \agent{Worker} two skills, \code{proof-refactoring.md} and \code{performance-optimization.md}, from {lean4-skills}~\citep{lean4-skills}. The first refactor under the new budget shows the effect: \href{https://github.com/YuanheZ/ErdosGraham/pull/447}{PR~\#447} replaced broad \code{nlinarith}/\code{simp} search in a peak-bound lemma with direct order arguments and hoisted shared sub-proofs, exactly that discipline.

\begin{table}[t]
  \centering
  \caption{Orchestration statistics for the three runs, computed from the agents' session histories. Cost is GPT-5.5 API-equivalent (\$5\,/\,\$0.50\,/\,\$30 per million input\,/\,cached-input\,/\,output tokens). Critical path is agent compute under parallelization; times are \texttt{hh:mm:ss}.}
  \label{tab:orchestration}
  \begin{tabular}{l|ccc}
    \toprule
    \textbf{Metric} & \textbf{Erd\H{o}s--Graham} & \textbf{ESS \#1196} & \textbf{\#164 \& \#1217} \\
    \midrule
    Rounds                       & 19 & 17 & 40 \\
    \agent{Worker}s launched     & 58 & 33 & 111 \\
    \agent{Refiner}s             & 7  & 6  & 25 \\
    Merged PRs                   & 53 & 32 & 93 \\
    \midrule
    Critical path (excl.\ CI wait) & 11:38:23 & 11:32:40 & 40:43:21 \\
    Aggregate agent active time    & 21:29:41 & 16:26:32 & 71:16:52 \\
    Stop-hook / CI wait time       & 07:15:10 & 02:15:33 & 07:05:26 \\
    Tool-call parsed wall time     & 10:16:25 & 06:16:40 & 49:08:03 \\
    Total tool calls               & 5{,}273 & 4{,}067 & 12{,}204 \\
    \midrule
    Total tokens                 & 308M & 245M & 796M \\
    GPT-5.5 API-equiv.\ cost     & \$257.17 & \$189.43 & \$623.54 \\
    \bottomrule
  \end{tabular}
\end{table}

\subsection{Case study: Erd\H{o}s--Graham}
\label{sec:cs-eg}

The Erd\H{o}s--Graham run shows where formalization effort concentrates. Its four target theorems were proven in $19$ rounds, yet every one of the $16$ refinement issues fell into just two families: Proposition~9's three-case tail bound, the analytic core of the general theorem ($8$ issues, spanning all seven refiner rounds), and the nested-interval covering lemma behind the construction theorem ($8$ issues, across $6$ rounds). No other node ever blocked. These are exactly the two results the baseline leaves unproven (\cref{sec:exp-baseline}).

The \agent{Refiner} converged each family by splitting the failed node into progressively finer sub-nodes that \agent{Worker}s discharged bottom-up. Most repairs corrected blueprint drift: a dyadic tail estimate that was false as written, its hypotheses letting the right-hand logarithmic factor vanish while the finite sum stayed positive so the claim collapsed to $1\le 0$ (\href{https://github.com/YuanheZ/ErdosGraham/issues/431}{issue~\#431}); a Lean index convention realigned to the paper's (\href{https://github.com/YuanheZ/ErdosGraham/issues/480}{issue~\#480}); and a strengthened statement that forced a complete proof to be downgraded to a placeholder rather than left unsound (\href{https://github.com/YuanheZ/ErdosGraham/issues/468}{issue~\#468}), the wholesale-replacement rule of \cref{sec:method-agents} in action.

The run also caught a genuine gap in the published proof. In Case~C of Proposition~9 the paper picks a Borel peak $R$ and the largest exponential failure $P<R$ beneath it, then uses $a_n\ge e^n$ throughout $[P+1,R+1]$. But the peak condition bounds only $a_{R+1}$, not $a_R$, and Case~C explicitly permits $a_R<e^R$: the chosen peak can itself be a failure, in which case $P=R$ and the block breaks at its lower end. A \agent{Worker} flagged the unjustified step (\href{https://github.com/YuanheZ/ErdosGraham/issues/469}{issue~\#469}) and the \agent{Refiner} repaired it by weakening the selection lemma to admit a failure at the peak (\href{https://github.com/YuanheZ/ErdosGraham/pull/473}{PR~\#473}), recovering the conclusion. This step is the analytic heart of the general theorem and exactly the kind of gap a machine-checked formalization is built to surface.

\subsection{Case study: \#1196}
\label{sec:cs-1196}

The \#1196 run shows the opposite pattern: one deep cascade rather than many independent defects, and no node refiled twice. Over five successive rounds the \agent{Refiner} drilled downward from the surface estimate to the genuinely missing core, inserting one deeper upstream lemma each round. The surface node needed the bound $-\zeta'(1+u)/\zeta(1+u)\le \log 2/(2^u-1)$; that reduced to monotonicity of the Dirichlet eta function; that reduced to stochastic domination of Gamma distributions in the shape parameter together with a Mellin representation of $\eta$. None of these are in Mathlib, so the harness built the entire probabilistic argument the paper compresses into a single sentence.

Once the chain reached the Mathlib-absent facts it converged with no further refinement: \agent{Worker}s proved the inserted leaves in dependency order, closing the original round-one defect seven rounds later from the bottom up. Five of the $6$ refiner rounds and $6$ of the eight issues targeted this one analytic spine, and the repairs added missing lemmas rather than rearranging existing ones, since the paper states in a line what a proof assistant needs a Gamma-coupling argument to establish. The eta-monotonicity step is precisely the blocker the baseline could not discharge (\cref{sec:exp-baseline}).

\subsection{Case study: \#164 and \#1217}
\label{sec:exp-casestudy}

The \code{Prim} run is the harness's hardest case and its sharpest illustration of the refinement loop. It ran 40 rounds and merged 93 pull requests (one \agent{Blueprinter}, 66 \agent{Worker}, and 26 \agent{Refiner}) while filing and closing 46 issues. It is by far the harness's largest run: its agents spent 71 hours of active compute, 41 of them along the critical path under parallelization, and \$624 in GPT-5.5-equivalent tokens (\cref{tab:orchestration}). The two targets behaved very differently: \#164, the Erd\H{o}s primitive set conjecture, was proven early, while \#1217, the divisibility-chain theorem, occupied the bulk of the run. The \agent{Refiner} ran in 24 of the 40 rounds; the last nine were refiner-free and completed the cleaned blueprint.

\paragraph{Drift dominates source-gaps.} The \agent{Refiner} classifies every illness area as \emph{drift} (the blueprint diverged from a sound source proof) or \emph{source-gap} (the paper's argument is itself incomplete); see \cref{sec:method-agents}. Its reports record 32 illness areas, split 26 to 6 in favor of drift. The drift defects are telling. A \agent{Worker}'s misformalized sub-invariance statement for \#164 was satisfied by the trivial identity kernel \texttt{P n m = [n=m]}, so it constrained nothing (\href{https://github.com/YuanheZ/Prim/issues/5}{issue~\#5}). Several statements asserted a real \code{tsum} or \code{limsup} of a divergent or unbounded quantity, which Lean silently totalizes to $0$; this hid the missing summability and forced a real-to-\code{ENNReal} reformulation of the chain-density definitions (issues~\href{https://github.com/YuanheZ/Prim/issues/2}{\#2}, \href{https://github.com/YuanheZ/Prim/issues/33}{\#33}, \href{https://github.com/YuanheZ/Prim/issues/102}{\#102}). The $6$ source-gaps are where the paper compresses: the invariant-weight asymptotic the authors call ``a routine calculation'' (\href{https://github.com/YuanheZ/Prim/issues/76}{issue~\#76}: ``not syntactic rewrites\dots\ substantive analytic obligations''), the occupation identity it derives by stating that ``an induction gives'' the result, and Mertens' theorem invoked as a black box (\href{https://github.com/YuanheZ/Prim/issues/127}{issue~\#127}).

\paragraph{Hard nodes converge over many rounds.} A few nodes were refiled and repaired repeatedly before converging (\cref{tab:recurrence}). The cluster behind \#1217 (an adjoint upward Markov chain, an occupation-measure identity, a uniform second-moment bound, and a reverse-Fatou extraction) dominated; its existence lemma \texttt{kernel-path-data-exists} was reopened in 10 distinct rounds. The recurrence came in two waves. Through round~21 the \agent{Refiner} pushed the construction upstream and bundled the analytic content (visit identity, second moment, reverse Fatou) into a single path-data package, so each downstream node could project what it needed (\href{https://github.com/YuanheZ/Prim/issues/69}{issue~\#69}: ``over-bundled for its position in the DAG''). At round~25 it reversed course and un-bundled the package to match the paper's derivation order; removing those fields invalidated four downstream projection proofs at once, each replaced wholesale with a placeholder. Round~26 then corrected the real-versus-\code{ENNReal} limsup semantics, and round~31 supplied the remaining analytic bridges: a measurability fix, a hit-count moment interface, and a Mertens estimate, the run's last source-gap. The node stabilized and the run closed.

\begin{table}[t]
  \centering
  \caption{The most-refiled blueprint nodes in the \code{Prim} run, with the span and number ($n$) of distinct \agent{Refiner} rounds that reopened each. Labels are abbreviated: the path-data, model, and reverse-Fatou nodes form the \#1217 adjoint-chain cluster (\code{mangoldt-adjoint-*}); the last is an EPS node (\code{eps-*}).}
  \label{tab:recurrence}
  \begin{tabular}{l|cc}
    \toprule
    \textbf{Blueprint node (abbreviated)} & \textbf{Target} & \textbf{Rounds ($n$)} \\
    \midrule
    \texttt{kernel-path-data-exists}        & \#1217 & 14--28 \ ($10$) \\
    \texttt{constructed-path-data-exists}   & \#1217 & 13--28 \ ($8$) \\
    \texttt{random-model-exists}            & \#1217 & 10--28 \ ($8$) \\
    \texttt{chain-density-selection}        & \#1217 & 9--26 \ ($7$) \\
    \texttt{reverse-fatou-path-extraction}  & \#1217 & 10--26 \ ($7$) \\
    \texttt{eps-modified-chain-subinvariant} & \#164  & 1--6 \ ($4$) \\
    \bottomrule
  \end{tabular}
\end{table}

\paragraph{The discipline holds.} The run produced 12 complete-proof downgrades across 7 rounds: when a parent's statement or type changed, the dependent proof was replaced wholesale, never partially edited (\cref{sec:method-agents}). The churn stayed safe because the downgrade is mechanical, decided by the Lean compiler, not negotiated by the agent.

\paragraph{Formalization expands the compressed steps.} Comparing the converged blueprint against the paper quantifies the gap. The \#1217 proof occupies about 62 lines of paper prose with no intermediate lemmas; the blueprint discharges it in roughly 84 nodes, a sixteen-fold expansion in lemma count. The single ``routine calculation'' for the von Mangoldt weight became about 14 explicit lemmas (a sum-integral interchange, a reciprocal-zeta derivative identification, and an integration-by-parts endpoint evaluation), and the phrase ``an induction gives'' became a 22-node construction of the probability space and its occupation measure. The expansion concentrates not in the number theory the authors wrote out carefully, but at the boundary where the paper defers to standard probability and analysis, exactly where a machine-checked proof cannot.

\subsection{Ablation: which design choices matter}
\label{sec:exp-ablation}

The Erd\H{o}s--Graham paper was first attempted with an earlier version of the harness, giving a controlled ablation on the same target. That run (PRs~\#46--\#429) differs from the one above in exactly two design choices, and it never finished: over roughly twelve days it filed $137$ issues and restarted its blueprint about eight times, ending with a \emph{larger} file ($12{,}910$ lines) that still carried $26$ \code{sorry} placeholders and none of the four target theorems proven.

\paragraph{The \agent{Refiner} must see the source.} In the earlier harness only the \agent{Blueprinter} ever read the source proof; the \agent{Refiner} saw only the blueprint DAG. Unable to separate blueprint drift from a genuine source gap, it answered blocked nodes by inventing upstream machinery the paper never contained, and the blueprint drifted further from the source each round. GitHub gave us enough observability to watch the run stall; we then intervened by hand, launching two human-in-the-loop passes that were given the source proof and tried to realign the blueprint (PRs~\href{https://github.com/YuanheZ/ErdosGraham/pull/405}{\#405}, \href{https://github.com/YuanheZ/ErdosGraham/pull/429}{\#429}). They diagnosed the damage exactly, finding that the blueprint had ``drifted toward early-prefix and moving-band chain-cover machinery as if those were the source construction'' (PR~\#429) and that an invented predicate was ``not the paper's Case (C)'' and ``forced the wrong trajectory'' (PR~\#405), but the drift was too deep to undo. That failure motivated the redesign and a fresh restart from scratch at \href{https://github.com/YuanheZ/ErdosGraham/pull/430}{PR~\#430}, in which the \agent{Refiner} reads the source proof and must classify every defect against it as drift or source gap. The current run closed all $16$ issues with $7$ source-aware repairs in a single pass.

\paragraph{The \agent{Worker} needs a principled stopping rule.} The earlier harness wrote its ``cannot formalize'' test directly into the \agent{Worker} spec as a budget of physical Lean lines: a node could be declared un-formalizable once its estimated proof exceeded a fixed line count. \agent{Worker}s gamed this rule, filing blocked-node issues that cited size rather than a real defect (``exceeds 1000 physical lines''), at nearly one block per \agent{Worker} PR. The current \agent{Worker} spec removes every size signal. Its four-phase audit permits an issue only on a concrete defect, e.g., a misformalized or false statement, a numerical counterexample, a Lean contradiction, or an invalid input, and explicitly forbids filing ``merely because the proof is substantial.'' A proof that is merely long, or that needs a helper the blueprint lacks, must instead be discharged inside the node's mechanically frozen edit region by growing a local refinement DAG, because ``a missing or out-of-order upstream helper is never a blocker.'' None of the new run's $16$ issues cite size, and every one names a real defect.

\paragraph{Takeaway.} The two changes are complementary and both necessary. A source-blind \agent{Refiner} turns repair into guesswork that compounds into drift; a budget-based \agent{Worker} turns ``cannot formalize'' into an escape hatch. Together they explain a twelve-day run that never converged; removing them turned the same paper into a fast, fully proven formalization.

\begin{table}[t]
  \centering
  \caption{Ablation on Erd\H{o}s--Graham: the same paper under the earlier harness (\agent{Refiner} blind to the source; \agent{Worker} governed by a physical-line budget) versus the current harness.}
  \label{tab:ablation}
  \begin{tabular}{l|cc}
    \toprule
    \textbf{Metric} & \textbf{Earlier harness} & \textbf{Current harness} \\
    \midrule
    Outcome                                  & stalled & \textbf{complete} \\
    Wall-clock                               & ${\sim}12$ days & ${\sim}3$ days \\
    Blueprint restarts                       & ${\sim}8$ & 1 \\
    Issues filed                             & 137 & 16 \\
    \quad citing a line budget               & 14 & 0 \\
    Source proof given to \agent{Refiner}    & no & yes \\
    Final \code{Main.lean}                   & 12{,}910\,ln, 26 \code{sorry} & 8{,}513\,ln, 0 \code{sorry} \\
    \bottomrule
  \end{tabular}
\end{table}

\subsection{Baseline: Aristotle}
\label{sec:exp-baseline}

We gave Aristotle the same inputs, the paper source and the target statements, and let it run to its own stopping point. It failed on both papers (\cref{tab:baseline}). On Erd\H{o}s--Graham it ran seven turns over $40$ hours and delivered $751$ lines with two \code{sorry}s. The two gaps are precisely the paper's deepest content: Proposition~9, the three-regime tail bound at the heart of the general theorem, and the nested-interval construction. Aristotle's own report calls both the paper's ``deepest mathematical content,'' beyond what it could automate. On \#1196 it ran four times over $24$ hours and delivered a $24$-line file whose single theorem is a \code{sorry}; its report identifies the missing piece as the flow inequality, equivalently the monotonicity of the Dirichlet eta function, and notes that the published proof ``has been formalized in Lean by Math Inc.,'' reading this as evidence that the task ``typically requires a dedicated team effort.'' \agentname closed exactly these gaps autonomously, proving every target with no \code{sorry}.
\begin{table}[ht]
  \centering
  \caption{Head-to-head with Aristotle on identical inputs. Aristotle ran ${>}40$\,h on Erd\H{o}s--Graham and ${>}24$\,h on \#1196 without eliminating its \code{sorry}s.}
  \label{tab:baseline}
  \begin{tabular}{l|cc|cc}
    \toprule
    & \multicolumn{2}{c|}{\textbf{Erd\H{o}s--Graham}} & \multicolumn{2}{c}{\textbf{ESS \#1196}} \\
    & Aristotle & \agentname & Aristotle & \agentname \\
    \midrule
    Targets proven         & 0\,/\,3 & 3\,/\,3 & 0\,/\,1 & 1\,/\,1 \\
    Lean lines delivered   & 751 & 8{,}513 & 24 & 3{,}988 \\
    Remaining \code{sorry} & 2 & 0 & 1 & 0 \\
    Outcome                & failed & \textbf{complete} & failed & \textbf{complete} \\
    \bottomrule
  \end{tabular}
\end{table}

\subsection{Failure case: the unit-distance disproof}
\label{sec:exp-failure}

We tried to formalize an OpenAI's recent disproof of the Erd\H{o}s unit-distance conjecture~\citep{openai2026unitdistance}. The disproof's clever geometric trick relies deep algebraic number theory, almost none of which exists in Mathlib. Our harness lets a proof stay incomplete only through explicit \code{sorry} placeholders and forbids assuming results as axioms per CI gate rejects, so with that theory missing the \agent{Blueprinter} had no honest foothold. The run instead faked the number theory: it modeled a number field as a dummy record and discharged the key step with placeholder values. This type-checks and passes CI, but proves nothing real. The run then stalled, stuck on the same node round after round, because the later geometric steps needed real objects the fake record could not supply, and the target was never reached. The lesson is a scope boundary, not about formalization capability: when the hardest part lives in prerequisites the library \textbf{significantly} lacks, the harness can organize the work but cannot fill the missing results since it is far away from the distribution of proof source.

\section{Conclusion}
\label{sec:conclusion}

We present \agentname, a long-running multi-agent harness that formalizes entire research papers into Lean~4. Treating long-horizon autoformalization as a problem of \emph{agent durability}, it decomposes a paper into an evolving proof DAG and contains faults behind four contract-scoped agents and a deterministic CI gate, turning one brittle multi-day run into many short, recoverable, parallel ones. On two 2026 papers spanning four Erd\H{o}s problems it formalizes all seven target theorems with no \code{sorry}, while a commercial agent baseline fails. \agentname is a step toward AI co-mathematicians whose long-running work stays legible, recoverable, and verifiable.

\section*{Acknowledgments and AI disclosure}
Y. Z. is supported by Warwick Chancellor’s International Scholarship and RIKEN-AIP Overseas Student Collaboration Program.
JDL acknowledges support of Open Philanthropy, NSF IIS 2107304, NSF CCF 2212262, ONR Young Investigator Award, NSF CAREER Award 2144994, and NSF CCF 2019844.
TS was partially supported by JSPS KAKENHI (24K02905) and JST CREST (JPMJCR2015).
YS was partially supported by the National Science Foundation under awards 2027737, 2113373, 2414918 and a gift from OpenAI.
This research is supported by the National Research Foundation, Singapore and the Ministry of Digital Development and Information under the AI Visiting Professorship Programme (award number AIVP-2024-004). Any opinions, findings and conclusions or recommendations expressed in this material are those of the author(s) and do not reflect the views of National Research Foundation, Singapore and the Ministry of Digital Development and Information.

ChatGPT was used to generate several images (e.g., \cref{fig-pipeline}) and proofread the paper.

\bibliographystyle{reference}
\bibliography{reference}

@misc{erdosGraham2026,
  title         = {Irrationality of rapidly converging series: a problem of {Erd\H{o}s} and {Graham}},
  author        = {Barreto, Kevin and Kang, Jiwon and Kim, Sang-hyun and Kova{\v{c}}, Vjekoslav and Zhang, Shengtong},
  year          = {2026},
  eprint        = {2601.21442},
  archivePrefix = {arXiv},
  primaryClass  = {math.NT},
  url           = {https://arxiv.org/abs/2601.21442}
}

@misc{primitiveChains2026,
  title         = {Primitive sets and von {Mangoldt} chains: {Erd\H{o}s} {Problem} \#1196 and beyond},
  author        = {Alexeev, Boris and Barreto, Kevin and Li, Yanyang and Lichtman, Jared Duker and Price, Liam and Shah, Jibran Iqbal and Tang, Quanyu and Tao, Terence},
  year          = {2026},
  eprint        = {2605.00301},
  archivePrefix = {arXiv},
  primaryClass  = {math.NT},
  url           = {https://arxiv.org/abs/2605.00301}
}

@misc{aletheia,
  title         = {Semi-Autonomous Mathematics Discovery with {Gemini}: A Case Study on the {Erd\H{o}s} Problems},
  author        = {Feng, Tony and Trinh, Trieu and Bingham, Garrett and Kang, Jiwon and Zhang, Shengtong and Kim, Sang-hyun and Barreto, Kevin and Schildkraut, Carl and Jung, Junehyuk and Seo, Jaehyeon and Pagano, Carlo and Chervonyi, Yuri and Hwang, Dawsen and Hou, Kaiying and Gukov, Sergei and Tsai, Cheng-Chiang and Choi, Hyunwoo and Jin, Youngbeom and Li, Wei-Yuan and Wu, Hao-An and Shiu, Ruey-An and Shih, Yu-Sheng and Le, Quoc V. and Luong, Thang},
  year          = {2026},
  eprint        = {2601.22401},
  archivePrefix = {arXiv},
  primaryClass  = {cs.AI},
  url           = {https://arxiv.org/abs/2601.22401}
}

@misc{bloom,
  title        = {{Erd\H{o}s} problems},
  author       = {Bloom, Thomas F.},
  year         = {2026},
  howpublished = {\url{https://www.erdosproblems.com}},
  note         = {Accessed 2026-05-30}
}

@inproceedings{zhang2026dagmath,
  title         = {{DAG-Math: Graph-of-Thought Guided Mathematical Reasoning in LLMs}},
  author        = {Zhang, Yuanhe and Kuzborskij, Ilja and Lee, Jason D. and Leng, Chenlei and Liu, Fanghui},
  booktitle     = {International Conference on Learning Representations (ICLR)},
  year          = {2026},
  eprint        = {2510.19842},
  archivePrefix = {arXiv},
  primaryClass  = {cs.LG},
  url           = {https://arxiv.org/abs/2510.19842}
}

@inproceedings{wu2022autoformalization,
  title         = {Autoformalization with Large Language Models},
  author        = {Wu, Yuhuai and Jiang, Albert Q. and Li, Wenda and Rabe, Markus N. and Staats, Charles and Jamnik, Mateja and Szegedy, Christian},
  booktitle     = {Advances in Neural Information Processing Systems (NeurIPS)},
  year          = {2022},
  eprint        = {2205.12615},
  archivePrefix = {arXiv},
  primaryClass  = {cs.LG},
  url           = {https://arxiv.org/abs/2205.12615}
}

@inproceedings{jiang2023dsp,
  title         = {Draft, Sketch, and Prove: Guiding Formal Theorem Provers with Informal Proofs},
  author        = {Jiang, Albert Q. and Welleck, Sean and Zhou, Jin Peng and Lacroix, Timoth{\'e}e and Liu, Jiacheng and Li, Wenda and Jamnik, Mateja and Lample, Guillaume and Wu, Yuhuai},
  booktitle     = {International Conference on Learning Representations (ICLR)},
  year          = {2023},
  eprint        = {2210.12283},
  archivePrefix = {arXiv},
  primaryClass  = {cs.AI},
  url           = {https://arxiv.org/abs/2210.12283}
}

@inproceedings{ying2024leanworkbook,
  title         = {Lean Workbook: A Large-Scale Lean Problem Set Formalized from Natural Language Math Problems},
  author        = {Ying, Huaiyuan and Wu, Zijian and Geng, Yihan and Wang, Jiayu and Lin, Dahua and Chen, Kai},
  booktitle     = {Advances in Neural Information Processing Systems (NeurIPS)},
  year          = {2024},
  eprint        = {2406.03847},
  archivePrefix = {arXiv},
  primaryClass  = {cs.AI},
  url           = {https://arxiv.org/abs/2406.03847}
}

@inproceedings{gao2025herald,
  title         = {Herald: A Natural Language Annotated {Lean}~4 Dataset},
  author        = {Gao, Guoxiong and Wang, Yutong and Jiang, Jiedong and Gao, Qi and Qin, Zihan and Xu, Tianyi and Dong, Bin},
  booktitle     = {International Conference on Learning Representations (ICLR)},
  year          = {2025},
  eprint        = {2410.10878},
  archivePrefix = {arXiv},
  primaryClass  = {cs.AI},
  url           = {https://arxiv.org/abs/2410.10878}
}

@misc{polu2020gptf,
  title         = {Generative Language Modeling for Automated Theorem Proving},
  author        = {Polu, Stanislas and Sutskever, Ilya},
  year          = {2020},
  eprint        = {2009.03393},
  archivePrefix = {arXiv},
  primaryClass  = {cs.LG},
  url           = {https://arxiv.org/abs/2009.03393}
}

@inproceedings{han2022pact,
  title         = {Proof Artifact Co-Training for Theorem Proving with Language Models},
  author        = {Han, Jesse Michael and Rute, Jason and Wu, Yuhuai and Ayers, Edward W. and Polu, Stanislas},
  booktitle     = {International Conference on Learning Representations (ICLR)},
  year          = {2022},
  eprint        = {2102.06203},
  archivePrefix = {arXiv},
  primaryClass  = {cs.LG},
  url           = {https://arxiv.org/abs/2102.06203}
}

@inproceedings{lample2022htps,
  title         = {{HyperTree} Proof Search for Neural Theorem Proving},
  author        = {Lample, Guillaume and Lacroix, Timoth{\'e}e and Lachaux, Marie-Anne and Rodriguez, Aur{\'e}lien and Hayat, Amaury and Lavril, Thibaut and Ebner, Gabriel and Martinet, Xavier},
  booktitle     = {Advances in Neural Information Processing Systems (NeurIPS)},
  year          = {2022},
  eprint        = {2205.11491},
  archivePrefix = {arXiv},
  primaryClass  = {cs.AI},
  url           = {https://arxiv.org/abs/2205.11491}
}

@inproceedings{yang2023leandojo,
  title         = {{LeanDojo}: Theorem Proving with Retrieval-Augmented Language Models},
  author        = {Yang, Kaiyu and Swope, Aidan M. and Gu, Alex and Chalamala, Rahul and Song, Peiyang and Yu, Shixing and Godil, Saad and Prenger, Ryan J. and Anandkumar, Anima},
  booktitle     = {Advances in Neural Information Processing Systems (NeurIPS)},
  year          = {2023},
  eprint        = {2306.15626},
  archivePrefix = {arXiv},
  primaryClass  = {cs.LG},
  url           = {https://arxiv.org/abs/2306.15626}
}

@inproceedings{lin2024leanstar,
  title         = {{Lean-STaR}: Learning to Interleave Thinking and Proving},
  author        = {Lin, Haohan and Sun, Zhiqing and Welleck, Sean and Yang, Yiming},
  booktitle     = {International Conference on Learning Representations (ICLR)},
  year          = {2025},
  eprint        = {2407.10040},
  archivePrefix = {arXiv},
  primaryClass  = {cs.AI},
  url           = {https://arxiv.org/abs/2407.10040}
}

@misc{xin2024deepseekproverv15,
  title         = {{DeepSeek-Prover-V1.5}: Harnessing Proof Assistant Feedback for Reinforcement Learning and Monte-Carlo Tree Search},
  author        = {Xin, Huajian and Ren, Z. Z. and Song, Junxiao and Shao, Zhihong and Zhao, Wanjia and Wang, Haocheng and Liu, Bo and Zhang, Liyue and Lu, Xuan and Du, Qiushi and Gao, Wenjun and Zhu, Qihao and Yang, Dejian and Gou, Zhibin and Wu, Z. F. and Luo, Fuli and Ruan, Chong},
  year          = {2024},
  eprint        = {2408.08152},
  archivePrefix = {arXiv},
  primaryClass  = {cs.CL},
  url           = {https://arxiv.org/abs/2408.08152}
}

@misc{ren2025deepseekproverv2,
  title         = {{DeepSeek-Prover-V2}: Advancing Formal Mathematical Reasoning via Reinforcement Learning for Subgoal Decomposition},
  author        = {Ren, Z. Z. and Shao, Zhihong and Song, Junxiao and Xin, Huajian and Wang, Haocheng and Zhao, Wanjia and Zhang, Liyue and Fu, Zhe and Zhu, Qihao and Yang, Dejian and Wu, Z. F. and Gou, Zhibin and Ma, Shirong and Tang, Hongxuan and Liu, Yuxuan and Gao, Wenjun and Guo, Daya and Ruan, Chong},
  year          = {2025},
  eprint        = {2504.21801},
  archivePrefix = {arXiv},
  primaryClass  = {cs.AI},
  url           = {https://arxiv.org/abs/2504.21801}
}

@misc{lin2025goedelproverv2,
  title         = {{Goedel-Prover-V2}: Scaling Formal Theorem Proving with Scaffolded Data Synthesis and Self-Correction},
  author        = {Lin, Yong and Tang, Shange and Lyu, Bohan and Yang, Ziran and Chung, Jui-Hui and Zhao, Haoyu and Jiang, Lai and Geng, Yihan and Ge, Jiawei and Sun, Jingruo and Wu, Jiayun and Gesi, Jiri and Lu, Ximing and Acuna, David and Yang, Kaiyu and Lin, Hongzhou and Choi, Yejin and Chen, Danqi and Arora, Sanjeev and Jin, Chi},
  year          = {2025},
  eprint        = {2508.03613},
  archivePrefix = {arXiv},
  primaryClass  = {cs.LG},
  url           = {https://arxiv.org/abs/2508.03613}
}

@misc{wang2025kiminaprover,
  title         = {{Kimina-Prover} Preview: Towards Large Formal Reasoning Models with Reinforcement Learning},
  author        = {Wang, Haiming and Unsal, Mert and Lin, Xiaohan and Baksys, Mantas and Liu, Junqi and Dos Santos, Marco and Sung, Flood and Vinyes, Marina and Ying, Zhenzhe and Zhu, Zekai and Lu, Jianqiao and de Saxc{\'e}, Hugues and others},
  year          = {2025},
  eprint        = {2504.11354},
  archivePrefix = {arXiv},
  primaryClass  = {cs.AI},
  url           = {https://arxiv.org/abs/2504.11354}
}

@misc{chen2025seedprover,
  title         = {{Seed-Prover}: Deep and Broad Reasoning for Automated Theorem Proving},
  author        = {Chen, Luoxin and Gu, Jinming and Huang, Liankai and Huang, Wenhao and Jiang, Zhicheng and Jie, Allan and Jin, Xiaoran and Jin, Xing and Li, Chenggang and Ma, Kaijing and Ren, Cheng and Shen, Jiawei and Shi, Wenlei and Sun, Tong and Sun, He and Wang, Jiahui and Wang, Siran and Wang, Zhihong and Wei, Chenrui and Wei, Shufa and others},
  year          = {2025},
  eprint        = {2507.23726},
  archivePrefix = {arXiv},
  primaryClass  = {cs.AI},
  url           = {https://arxiv.org/abs/2507.23726}
}

@article{hubert2026alphaproof,
  title   = {Olympiad-level formal mathematical reasoning with reinforcement learning},
  author  = {Hubert, Thomas and Mehta, Rishi and Sartran, Laurent and Horv{\'a}th, Mikl{\'o}s Z. and {\v{Z}}u{\v{z}}i{\'c}, Goran and Wieser, Eric and Huang, Aja and Schrittwieser, Julian and Schroecker, Yannick and Masoom, Hussain and Bertolli, Ottavia and Zahavy, Tom and Mandhane, Amol and Yung, Jessica and Beloshapka, Iuliya and Ibarz, Borja and Veeriah, Vivek and Yu, Lei and Nash, Oliver and Lezeau, Paul and Mercuri, Salvatore and S{\"o}nne, Calle and Mehta, Bhavik and Davies, Alex and Zheng, Daniel and Pedregosa, Fabian and Li, Yin and von Glehn, Ingrid and Rowland, Mark and Albanie, Samuel and Velingker, Ameya and Schmitt, Simon and Lockhart, Edward and Hughes, Edward and Michalewski, Henryk and Sonnerat, Nicolas and Hassabis, Demis and Kohli, Pushmeet and Silver, David},
  journal = {Nature},
  volume  = {651},
  pages   = {607--613},
  year    = {2026},
  doi     = {10.1038/s41586-025-09833-y},
  url     = {https://doi.org/10.1038/s41586-025-09833-y}
}

@article{trinh2024alphageometry,
  title   = {Solving olympiad geometry without human demonstrations},
  author  = {Trinh, Trieu H. and Wu, Yuhuai and Le, Quoc V. and He, He and Luong, Thang},
  journal = {Nature},
  volume  = {625},
  number  = {7995},
  pages   = {476--482},
  year    = {2024},
  doi     = {10.1038/s41586-023-06747-5},
  url     = {https://doi.org/10.1038/s41586-023-06747-5}
}

@misc{novikov2025alphaevolve,
  title         = {{AlphaEvolve}: A Coding Agent for Scientific and Algorithmic Discovery},
  author        = {Novikov, Alexander and V{\~u}, Ng{\^a}n and Eisenberger, Marvin and Dupont, Emilien and Huang, Po-Sen and Wagner, Adam Zsolt and Shirobokov, Sergey and Kozlovskii, Borislav and Ruiz, Francisco J. R. and Mehrabian, Abbas and Kumar, M. Pawan and See, Abigail and Chaudhuri, Swarat and Holland, George and Davies, Alex and Nowozin, Sebastian and Kohli, Pushmeet and Balog, Matej},
  year          = {2025},
  eprint        = {2506.13131},
  archivePrefix = {arXiv},
  primaryClass  = {cs.NE},
  url           = {https://arxiv.org/abs/2506.13131}
}

@misc{glazer2024frontiermath,
  title         = {{FrontierMath}: A Benchmark for Evaluating Advanced Mathematical Reasoning in {AI}},
  author        = {Glazer, Elliot and Erdil, Ege and Besiroglu, Tamay and Chicharro, Diego and Chen, Evan and Gunning, Alex and Olsson, Caroline Falkman and Denain, Jean-Stanislas and Ho, Anson and de Oliveira Santos, Emily and J{\"a}rviniemi, Olli and Barnett, Matthew and Sandler, Robert and Vrzala, Matej and Sevilla, Jaime and Ren, Qiuyu and Pratt, Elizabeth and Levine, Lionel and Barkley, Grant and Stewart, Natalie and Grechuk, Bogdan and Grechuk, Tetiana and Enugandla, Shreepranav Varma and Wildon, Mark},
  year          = {2024},
  eprint        = {2411.04872},
  archivePrefix = {arXiv},
  primaryClass  = {cs.AI},
  url           = {https://arxiv.org/abs/2411.04872}
}

@misc{bubeck2025gpt5science,
  title         = {Early Science Acceleration Experiments with {GPT-5}},
  author        = {Bubeck, S{\'e}bastien and Coester, Christian and Eldan, Ronen and Gowers, Timothy and Lee, Yin Tat and Lupsasca, Alexandru and Sawhney, Mehtaab and Scherrer, Robert and Sellke, Mark and Spears, Brian K. and Unutmaz, Derya and Weil, Kevin and Yin, Steven and Zhivotovskiy, Nikita},
  year          = {2025},
  eprint        = {2511.16072},
  archivePrefix = {arXiv},
  primaryClass  = {cs.AI},
  url           = {https://arxiv.org/abs/2511.16072}
}

@article{tao2025machineassisted,
  title   = {Machine-Assisted Proof},
  author  = {Tao, Terence},
  journal = {Notices of the American Mathematical Society},
  volume  = {72},
  number  = {1},
  pages   = {6--13},
  year    = {2025},
  doi     = {10.1090/noti3041},
  url     = {https://doi.org/10.1090/noti3041}
}

@misc{tsoukalas2026formalproofsearch,
  title         = {Advancing Mathematics Research with {AI}-Driven Formal Proof Search},
  author        = {Tsoukalas, George and Kovsharov, Anton and Shirobokov, Sergey and Surina, Anja and Firsching, Moritz and B{\'e}rczi, Gergely and Ruiz, Francisco J. R. and Suggala, Arun and Wagner, Adam Zsolt and Wieser, Eric and Yu, Lei and Huang, Aja and Horv{\'a}th, Mikl{\'o}s Z. and Ferrauiolo, Andrew and Michalewski, Henryk and Grosu, Codrut and Hubert, Thomas and Balog, Matej and Kohli, Pushmeet and Chaudhuri, Swarat},
  year          = {2026},
  eprint        = {2605.22763},
  archivePrefix = {arXiv},
  primaryClass  = {cs.AI},
  url           = {https://arxiv.org/abs/2605.22763}
}

@article{aristotle,
  title={{Aristotle: IMO-level Automated Theorem Proving}},
  author={Achim, Tudor and Best, Alex and Bietti, Alberto and Der, Kevin and F{\'e}d{\'e}rico, Math{\"\i}s and Gukov, Sergei and Halpern-Leistner, Daniel and Henningsgard, Kirsten and Kudryashov, Yury and Meiburg, Alexander and others},
  journal={arXiv preprint arXiv:2510.01346},
  year={2025}
}

@misc{gauss1196,
  title        = {{Erdos1196}: A {Lean} formalization of {Erd\H{o}s} {Problem} \#1196},
  author       = {{Math, Inc.}},
  year         = {2026},
  howpublished = {\url{https://github.com/math-inc/Erdos1196}},
  note         = {Formalized by the Gauss autoformalization agent. Accessed 2026-05-30}
}

@misc{massot2020leanblueprint,
  title        = {leanblueprint: A {plasTeX} plugin to build formalization blueprints for {Lean}},
  author       = {Massot, Patrick},
  year         = {2020},
  howpublished = {\url{https://github.com/PatrickMassot/leanblueprint}},
  note         = {Accessed 2026-05-30}
}

@misc{zhu2026leanarchitect,
  title         = {{LeanArchitect}: Automating Blueprint Generation for Humans and {AI}},
  author        = {Zhu, Thomas and Monticone, Pietro and Avigad, Jeremy and Welleck, Sean},
  year          = {2026},
  eprint        = {2601.22554},
  archivePrefix = {arXiv},
  primaryClass  = {cs.LO},
  url           = {https://arxiv.org/abs/2601.22554}
}

@misc{pfr2023,
  title        = {Formalization of the Polynomial {Freiman-Ruzsa} Conjecture of {Marton}},
  author       = {Dillies, Ya{\"e}l and Tao, Terence and others},
  year         = {2023},
  howpublished = {\url{https://github.com/teorth/pfr}},
  note         = {{Lean}~4 formalization project. Accessed 2026-05-30}
}

@article{scholze2022liquid,
  title   = {Liquid Tensor Experiment},
  author  = {Scholze, Peter},
  journal = {Experimental Mathematics},
  volume  = {31},
  number  = {2},
  pages   = {349--354},
  year    = {2022},
  doi     = {10.1080/10586458.2021.1926016},
  url     = {https://doi.org/10.1080/10586458.2021.1926016}
}

@misc{buzzard2024flt,
  title        = {The {Fermat's} {Last} {Theorem} Project},
  author       = {Buzzard, Kevin and others},
  year         = {2024},
  howpublished = {\url{https://github.com/ImperialCollegeLondon/FLT}},
  note         = {{Lean}~4 formalization project, Imperial College London. Accessed 2026-05-30}
}

@misc{becker2024carleson,
  title         = {A Blueprint for the Formalization of {Carleson's} Theorem on Convergence of {Fourier} Series},
  author        = {Becker, Lars and de Frutos-Fern{\'a}ndez, Mar{\'i}a In{\'e}s and Diedering, Leo and van Doorn, Floris and Gou{\"e}zel, S{\'e}bastien and Jamneshan, Asgar and Karunus, Evgenia and van de Meent, Edward and Monticone, Pietro and Mulder-Sohn, Jasper and Portegies, Jim and Roos, Joris and Rothgang, Michael and Srivastava, Rajula and Sundstrom, James and Tan, Jeremy and Thiele, Christoph},
  year          = {2024},
  eprint        = {2405.06423},
  archivePrefix = {arXiv},
  primaryClass  = {math.CA},
  url           = {https://arxiv.org/abs/2405.06423}
}

@misc{bolan2025equational,
  title         = {The {Equational} {Theories} {Project}: Advancing Collaborative Mathematical Research at Scale},
  author        = {Bolan, Matthew and Breitner, Joachim and Brox, Jose and Carlini, Nicholas and Carneiro, Mario and van Doorn, Floris and Dvo{\v{r}}{\'a}k, Martin and Goens, Andr{\'e}s and Hill, Aaron and Husum, Harald and Ibarra Mejia, Hern{\'a}n and Kocsis, Zoltan A. and Le Floch, Bruno and Livne Bar-on, Amir and Luccioli, Lorenzo and McNeil, Douglas and Meiburg, Alex and Monticone, Pietro and Nielsen, Pace P. and Osazuwa, Emmanuel Osalotioman and Paolini, Giovanni and Petracci, Marco and Reinke, Bernhard and Renshaw, David and Rossel, Marcus and Roux, Cody and Scanvic, J{\'e}r{\'e}my and Srinivas, Shreyas and Tadipatri, Anand Rao and Tao, Terence and Tsyrklevich, Vlad and Vaquerizo-Villar, Fernando and Weber, Daniel and Zheng, Fan},
  year          = {2025},
  eprint        = {2512.07087},
  archivePrefix = {arXiv},
  primaryClass  = {cs.LO},
  url           = {https://arxiv.org/abs/2512.07087}
}

@misc{tao2026threecomponents,
  author       = {Tao, Terence},
  title        = {The three components of problem solving: proof generation, proof verification, and proof digestion},
  year         = {2026},
  month        = April,
  day          = {22},
  howpublished = {Mathstodon post},
  url          = {https://mathstodon.xyz/@tao/116450581967483825},
  note         = {Accessed: 2026-06-01}
}

@article{zheng2026comath,
  title={{AI Co-Mathematician: Accelerating Mathematicians with Agentic AI}},
  author={Zheng, Daniel and von Glehn, Ingrid and Zwols, Yori and Beloshapka, Iuliya and Buesing, Lars and Roy, Daniel M and Wattenberg, Martin and Georgiev, Bogdan and Schmidt, Tatiana and Cowie, Andrew and others},
  journal={arXiv preprint arXiv:2605.06651},
  year={2026}
}

@article{hariharan2026gauss,
  title={A Milestone in Formalization: The Sphere Packing Problem in Dimension 8},
  author={Hariharan, Sidharth and Birkbeck, Christopher and Lee, Seewoo and Ma, Ho Kiu Gareth and Mehta, Bhavik and Poiroux, Auguste and Viazovska, Maryna},
  journal={arXiv preprint arXiv:2604.23468},
  year={2026}
}

@misc{axiommath2025,
  author       = {{Axiom Math}},
  title        = {{Axiom Math}},
  year         = {2025},
  howpublished = {Website},
  url          = {https://axiommath.ai/},
  note         = {Accessed: 2026-06-01}
}

@article{sonoda2025lean,
  title={Lean formalization of generalization error bound by rademacher complexity},
  author={Sonoda, Sho and Kasaura, Kazumi and Mizuno, Yuma and Tsukamoto, Kei and Onda, Naoto},
  journal={arXiv preprint arXiv:2503.19605},
  year={2025}
}

@inproceedings{
zhang2026statistical,
title={{AI4SLT: Empirical Processes in Lean 4 for Formal Statistical Learning Theory}},
author={Zhang, Yuanhe and Lee, Jason D and Liu, Fanghui},
booktitle={Forty-third International Conference on Machine Learning},
year={2026},
url={https://openreview.net/forum?id=dfqmQ9WhCP}
}

@article{zhang2025towards,
  title={Towards formalizing reinforcement learning theory},
  author={Zhang, Shangtong},
  journal={arXiv preprint arXiv:2511.03618},
  year={2025}
}

@misc{lean4-skills,
  author = {Cameron Freer},
  title = {{Lean 4 Skills: Theorem proving skill and workflow pack for AI coding agents}},
  url = {https://github.com/cameronfreer/lean4-skills},
  month = oct,
  year = {2025}
}

@misc{openai2026unitdistance,
  title        = {An {OpenAI} Model Has Disproved a Central Conjecture in Discrete Geometry},
  author       = {{OpenAI}},
  year         = {2026},
  howpublished = {\url{https://openai.com/index/model-disproves-discrete-geometry-conjecture/}},
  note         = {Accessed 2026-06-01}
}

@article{wang2026m2f,
  title={M2F: Automated Formalization of Mathematical Literature at Scale},
  author={Wang, Zichen and Ma, Wanli and Ming, Zhenyu and Zhang, Gong and Yuan, Kun and Wen, Zaiwen},
  journal={arXiv preprint arXiv:2602.17016},
  year={2026}
}

@article{gloeckle2026automatic,
  title={Automatic textbook formalization},
  author={Gloeckle, Fabian and Rammal, Ahmad and Arnal, Charles and Munos, Remi and Cabannes, Vivien and Synnaeve, Gabriel and Hayat, Amaury},
  journal={arXiv preprint arXiv:2604.03071},
  year={2026}
}

\newpage
\appendix
\tableofcontents
\newpage

\section{Agent Knowledge-Store Layouts}
\label{app:agent-layouts}

Each agent runs in an isolated git worktree whose knowledge store is fixed before launch; it reads nothing outside it. \cref{tree:blueprinter,tree:reviewer,tree:worker,tree:refiner} give the four layouts, and the file roles are as follows.

\paragraph{Runtime.} \code{AGENTS.md} is each agent's charter: its purpose, hard boundaries, inputs, and ordered workflow. \code{.codex/config.toml} fixes the model, the read-only sandbox, and the exact MCP tools the agent may call; \code{rules/default.rules} (shared verbatim) forbids \code{git}, \code{gh}, and network commands; and \code{hooks/ralph\_wiggum\_stop.py} is the stop hook that blocks exit until \code{delivery.yml} records a merged PR, polling CI and re-injecting failed-job logs (the \agent{Worker}'s variant also accepts an issue as terminal). The \agent{Target-Reviewer} is single-shot and carries no hook.

\paragraph{Inputs and state.} \code{docs/inputs.yml} lists the run's file paths and \code{owner}/\code{repo}/\code{branch}; it carries \code{proof\_file}, the raw source proof, only for the \agent{Blueprinter} and \agent{Refiner}. \code{docs/delivery.yml} is the terminal-delivery record the stop hook validates.
Each agent maintains a tiny \code{state.md} with summaries per phase and per delivery path. Statuses are classified into \code{none}, \code{in-progress}, \code{pass}, \code{fail}, \code{complete}. The recovery rules after an auto-compaction are explicit and ordered: if delivery is \code{complete}, exit; if exactly one row is \code{in-progress}, resume it; if a phase is \code{fail}, jump to issue delivery; if the last execution phase is \code{pass}, jump to PR delivery; otherwise start the earliest \code{none} phase. State never overrides the agent's \code{AGENTS.md}, the phase contracts, or the Lean diagnostics. The combination guarantees that no compaction re-runs a phase that already passed, and no compaction skips a phase that already failed.

\paragraph{Contracts.} \code{contracts/blueprint-format.md} is the \code{@[blueprint]} formatting contract the CI enforces (a placeholder-only variant for the \agent{Blueprinter}, a complete-proof variant for the \agent{Worker} and \agent{Refiner}); \code{contracts/latex-quality.md} sets the prose rules with an agent-specific honesty-about-source clause. The \agent{Worker} alone also carries \code{contracts/edit-constraints.md}, which partitions the file into frozen and editable line spans enforced by the \code{apply-patch} server, and \code{contracts/local-refinement.md}, which governs the local helper-lemma DAG it may grow before its target.

\paragraph{Execution phases.} \code{docs/exec-phases/} holds one ordered \code{TASK.md} per phase: \code{understand}/\\\code{grounding}/\code{draft} for the \agent{Blueprinter}, a single audit phase for the \agent{Target-Reviewer},\\ \code{misformalization}/\code{numeric}/\code{polish}/\code{formalization} for the \agent{Worker} (\cref{fig:worker-workflow}), and \code{scope}/\\\code{refine} for the \agent{Refiner}, whose \code{scope} phase classifies each illness area against \code{proof\_file} as drift or source gap.

\paragraph{Delivery and references.} \code{docs/deliver/pr.md} and \code{issue.md} are the PR and issue templates; the \agent{Target-Reviewer} files only issues, the \agent{Worker} may do either, and the \agent{Blueprinter} and \agent{Refiner} only open PRs. \code{grounding/SKILL.md} and its worked examples drive just-in-time Mathlib retrieval; the remaining \code{references/} are read-on-demand know-how, including the \agent{Blueprinter}'s \code{decomposition.md} (the node-decomposition rubric) and \code{reframing.md}, the \agent{Worker}'s \code{formalization-style.md},\\ \code{lean-lsp-tools.md}, \code{proof-refactoring.md}, and \code{performance-optimization.md}, and the shared \code{numeric-tools.md}, \code{compute.md}, and \code{pdf-reading.md}.

\begin{figure}[ht]
\dirtree{%
.1 Blueprinter/.
.2 AGENTS.md.
.2 .codex/.
.3 config.toml.
.3 rules/default.rules.
.3 hooks/ralph\_wiggum\_stop.py.
.2 docs/.
.3 inputs.yml.
.3 state.md.
.3 delivery.yml.
.3 contracts/.
.4 blueprint-format.md.
.4 latex-quality.md.
.3 deliver/.
.4 pr.md.
.3 exec-phases/.
.4 understand/TASK.md.
.4 grounding/TASK.md.
.4 draft/TASK.md.
.3 references/.
.4 decomposition.md.
.4 reframing.md.
.4 discovery-example.md.
.4 api-example.md.
.4 pdf-reading.md.
}
\caption{Knowledge store of the \agent{Blueprinter}.}
\label{tree:blueprinter}
\end{figure}

\begin{figure}[ht]
\dirtree{%
.1 Target-Reviewer/.
.2 AGENTS.md.
.2 .codex/.
.3 config.toml.
.3 rules/default.rules.
.2 docs/.
.3 inputs.yml.
.3 deliver/.
.4 issue.md.
.3 exec-phase/.
.4 TASK.md.
.3 grounding/.
.4 SKILL.md.
.4 example/api-example.md.
}
\caption{Knowledge store of the \agent{Target-Reviewer}. It carries no \code{proof\_file} and no PR delivery path: it audits and files issues only.}
\label{tree:reviewer}
\end{figure}

\begin{figure}[ht]
\dirtree{%
.1 Worker/.
.2 AGENTS.md.
.2 .codex/.
.3 config.toml.
.3 rules/default.rules.
.3 hooks/ralph\_wiggum\_stop.py.
.2 docs/.
.3 inputs.yml.
.3 state.md.
.3 delivery.yml.
.3 contracts/.
.4 blueprint-format.md.
.4 edit-constraints.md.
.4 local-refinement.md.
.4 latex-quality.md.
.3 deliver/.
.4 issue.md.
.4 pr.md.
.3 exec-phases/.
.4 misformalization/TASK.md.
.4 numeric/TASK.md.
.4 polish/TASK.md.
.4 formalization/TASK.md.
.3 grounding/.
.4 SKILL.md.
.4 examples/api-example.md.
.4 examples/discovery-example.md.
.3 references/.
.4 formalization-style.md.
.4 lean-lsp-tools.md.
.4 proof-refactoring.md.
.4 performance-optimization.md.
.4 numeric-tools.md.
.4 compute.md.
}
\caption{Knowledge store of the per-node \agent{Worker}. The four \code{exec-phases} mirror \cref{fig:worker-workflow}; \code{edit-constraints.md} and \code{local-refinement.md} instruct the mechanically enforced editable region.}
\label{tree:worker}
\end{figure}

\begin{figure}[ht]
\dirtree{%
.1 Refiner/.
.2 AGENTS.md.
.2 .codex/.
.3 config.toml.
.3 rules/default.rules.
.3 hooks/ralph\_wiggum\_stop.py.
.2 docs/.
.3 inputs.yml.
.3 state.md.
.3 delivery.yml.
.3 contracts/.
.4 blueprint-format.md.
.4 latex-quality.md.
.3 deliver/.
.4 pr.md.
.3 exec-phases/.
.4 scope/TASK.md.
.4 refine/TASK.md.
.3 grounding/.
.4 SKILL.md.
.4 examples/api-example.md.
.4 examples/discovery-example.md.
.3 references/.
.4 compute.md.
.4 numeric-tools.md.
.4 pdf-reading.md.
}
\caption{Knowledge store of the \agent{Refiner}. Its \code{docs/inputs.yml} carries a \code{proof\_file} (the source proof) and an \code{issues\_file}; its two \code{exec-phases} are \code{scope} (locate the illness area) and \code{refine}.}
\label{tree:refiner}
\end{figure}

\end{document}